\documentclass[lettersize,journal]{IEEEtran}
\usepackage{amsmath,amsfonts}
\usepackage{algorithmic}
\usepackage{algorithm}
\usepackage{array}
\usepackage[caption=false,font=normalsize,labelfont=sf,textfont=sf]{subfig}
\usepackage{textcomp}
\usepackage{stfloats}
\usepackage{url}
\usepackage{verbatim}
\usepackage{graphicx}
\usepackage{cite}
\usepackage{bm}
\usepackage[table]{xcolor} 
\usepackage{colortbl} 
\usepackage{array}
\usepackage{tcolorbox}

\newcommand{\std}[1]{{\tiny($\pm$#1)}}

\usepackage[utf8]{inputenc}
\usepackage[T1]{fontenc}
\usepackage{etoolbox}
\usepackage{bbm}
\usepackage{svg}

\usepackage{amsmath}
\usepackage{siunitx}
\usepackage{mathptmx}
\usepackage{mathrsfs}
\usepackage{hyperref}
\usepackage{pifont}
\usepackage{booktabs}
\usepackage{multirow}
\usepackage{threeparttable}
\usepackage{diagbox}
\usepackage{tabularx}
\usepackage{placeins} 
\usepackage{amsthm,amsmath,amssymb}
\usepackage{booktabs}
\usepackage{multirow}
\usepackage{mathrsfs}
\usepackage{adjustbox}
\usepackage{enumerate}
\usepackage{mathrsfs}
\usepackage{makecell}
\usepackage{diagbox}
\usepackage{algorithm}
\usepackage{algorithmic}

\DeclareMathAlphabet{\mathcal}{OMS}{cmsy}{m}{n}
\DeclareSymbolFont{largesymbols}{OMX}{cmex}{m}{n}
\usepackage {threeparttable}
\hypersetup{hidelinks}
\begin{document}

\title{DRAN: A Distribution and Relation Adaptive Network for Spatio-temporal Forecasting}

\author{
Xiaobei Zou, Luolin Xiong, Kexuan Zhang, Cesare Alippi,~\IEEEmembership{Fellow,~IEEE}, Yang Tang,~\IEEEmembership{Fellow,~IEEE}
\thanks{
This work was supported by the National Natural Science Foundation of China under Grants 62293502, 62233005, U2441245, and 62603271; in part by the Shanghai Municipal Special Program for Basic Research on General AI Foundation Models under Grant 2025SHZDZX026D01; in part by the Shanghai Baiyulan Talent Program Pujiang Project under Grant 25PJD028; and in part by the China Postdoctoral Science Foundation under Grants BX2026282 and 2025M781640. The computations in this research were performed using the CFFF platform of Fudan University. \textit{(Corresponding author: Yang Tang.)}}
\thanks{Xiaobei Zou, Luolin Xiong, Kexuan Zhang and Yang Tang are with the Key Laboratory of Smart Manufacturing in Energy Chemical Process, Ministry of Education, East China University of Science and Technology, Shanghai 200237, China (e-mail: xbeizou@gmail.com; xiongluolin@gmail.com; kexuanzhang123@gmail.com; yangtang@ecust.edu.cn).}
\thanks{Cesare Alippi is with the Faculty of Informatics, Università della Svizzera italiana, 69000 Lugano, Switzerland, and also with the Department of Electronics, Information and Bioengineering, Politecnico di Milano, 20133 Milan,
Italy (e-mail: alippi@elet.polimi.it).
}
}

\markboth{IEEE TRANSACTIONS ON CYBERNETICS}%
{Shell \MakeLowercase{\textit{et al.}}: A Sample Article Using IEEEtran.cls for IEEE Journals}

\IEEEpubid{}

\maketitle

\begin{abstract}
Spatio-temporal forecasting remains challenging under non-stationary environments because both data distributions and spatial relations evolve over time. Temporal normalization and de-normalization are widely used to mitigate distribution shifts, but they may distort inter-node relationships and thereby impair spatial dependency modeling.
To address these issues, we propose the \underline{D}istribution and \underline{R}elation \underline{A}daptive \underline{N}etwork (DRAN) for spatio-temporal forecasting. 
DRAN incorporates a Spatial Factor Learner (SFL) module, which enables effective normalization and de-normalization while preserving spatial dependencies in spatio-temporal systems.
To model evolving spatial interactions, DRAN further proposes the Dynamic-Static Fusion Learner (DSFL) module. DSFL decomposes features into static and dynamic components and adaptively fuses them according to input variability.
Experiments on six benchmark datasets show that DRAN outperforms state-of-the-art baselines. Additional analyses demonstrate that SFL consistently reduces spatial-relation distortion across multiple normalization schemes, whereas DSFL captures complementary static and dynamic dependencies and adjusts their contributions according to temporal variability.
\end{abstract}

\begin{IEEEkeywords}
Spatio-temporal forecasting, graph neural network, distribution adaptation, adaptive network.
\end{IEEEkeywords}

\section{Introduction}
\IEEEPARstart{S}{patio-temporal} systems, characterized by intricate spatial interactions among sensors (nodes) and complex temporal dynamics, are prevalent across a wide range of fields, including physics \cite{10491369}, meteorology \cite{9457154}, power grids \cite{ga2022, xiong2022two} and transportation \cite{9983531}. These systems often involve a large number of nodes, with interactions that vary over time \cite{wu2022survey, ji2023signal}. The high complexity and temporal variability inherent in such systems (time variance) make it challenging to effectively manage future developments and support informed decision-making, thereby necessitating more accurate spatio-temporal prediction methods \cite{9764831}. \par
In spatio-temporal forecasting, historical time series data associated with spatially distributed nodes are used to predict future observations \cite{Cini_Alippi_2023}. Although numerous effective methods, particularly deep learning-based approaches, have been proposed \cite{tang2020introduction, samsgl}, accurate forecasting remains a challenge due to the time-varying nature of interactions between the environment and the system.\par
\begin{figure}[!t]
\centering
\includegraphics[width=\linewidth]{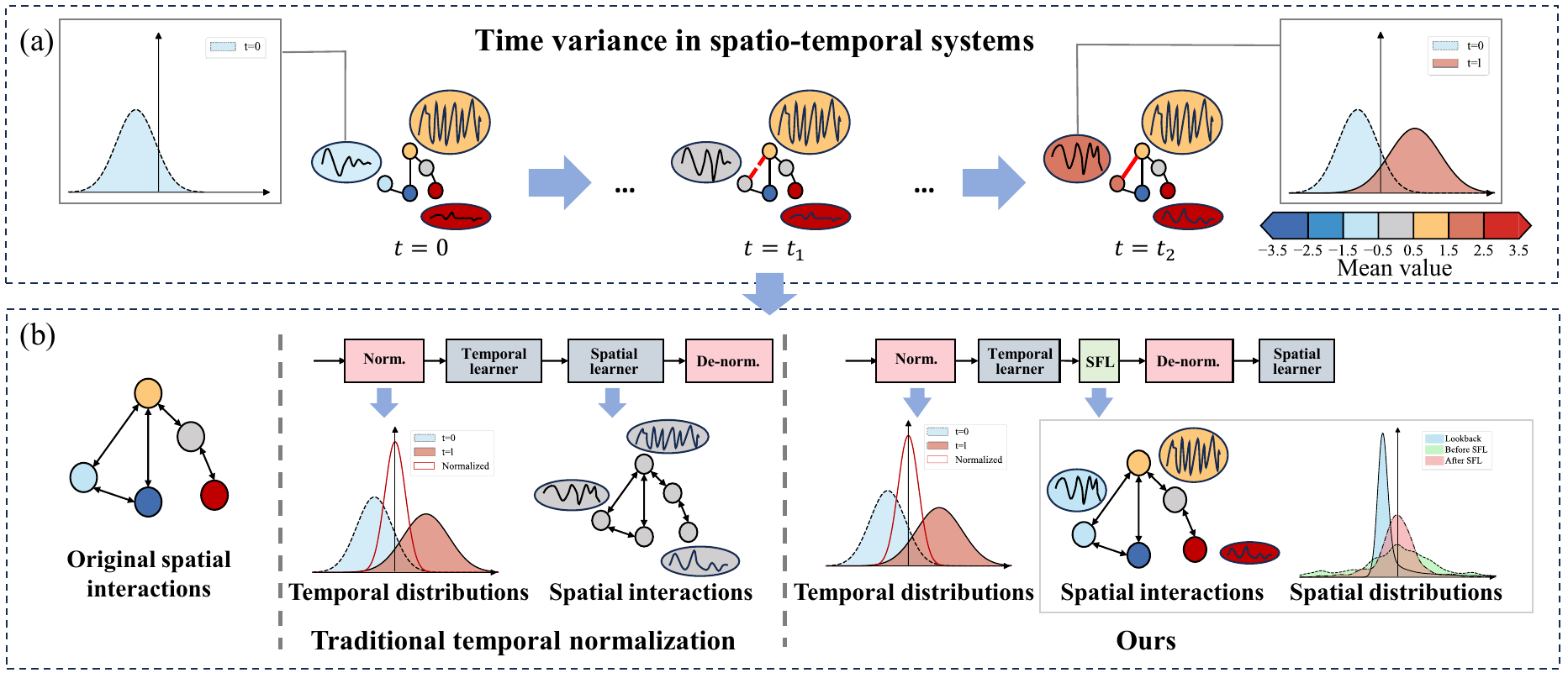}
\caption{Time variance in spatio-temporal systems and corresponding distribution adaptation strategies. (a) Temporal variations of a spatio-temporal system, including shifts in temporal distributions and changes in node connectivity. The color of each node represents the mean value of the node variables (consistent with the colorbar), while the vertical size of each ellipse indicates the variance. (b) Comparison between conventional normalization and de-normalization strategies for distribution adaptation, which may disrupt the effectiveness of spatial relational learning, and the proposed SFL module for preserving spatial distribution structures. ``Norm.'' and ``De-norm.'' denote normalization and de-normalization, respectively.}
\label{dis_case}
\end{figure}

As illustrated in Fig. \ref{dis_case}, spatio-temporal systems evolve over time, exhibiting variations in temporal distributions and spatial connections among nodes.
In many spatio-temporal forecasting models, temporal and spatial features are progressively learned. Normalization and de-normalization strategies \cite{kim2021reversible, liu2022non, fan2023dish} are often incorporated to mitigate distribution shifts, where all nodes are rescaled to a unified mean and variance. However, in spatio-temporal contexts where nodes are spatially connected and mutually influenced, accurate modeling of spatial interactions is equally important for reliable forecasting. As is shown in Fig. \ref{dis_case} (b), 
temporal normalization rescales nodes to the same mean and variance level (grey), causing the learned spatial interactions to deviate from the original spatial relationships.
Therefore, there is a clear need for normalization approaches that maintain spatial consistency while adapting to temporal variation in spatio-temporal forecasting tasks. \par
The connections among nodes may gradually evolve over time (Fig. \ref{dis_case}(a)), with edges being preserved, newly formed, or removed. To capture the time-varying nature of spatial relationships, various approaches have been proposed to learn static \cite{zhao2019t, yu2018spatio, seo2018structured} and dynamic spatial structures \cite{liu2023spatio}. To enhance spatial representation learning, decomposition-based methods have also been introduced to enable hierarchical modeling, separating spatial information according to spectral frequency \cite{decomp_spec, decomp_freq}, spatial scales \cite{fang2023learning}, or edge direction \cite{decomp_direct}, and subsequently integrating them through learnable weights or fusion mechanisms \cite{li2023dynamic, fang2023learning}.
However, existing models often neglect that the interplay among decomposed components may also vary over time. For instance, in traffic systems, dynamic patterns tend to dominate during peak hours, whereas static and stable structures become more prominent during low-demand periods.
Consequently, there is a pressing need to develop methods capable of decomposing and adaptively fusing both static and dynamic spatial representations based on dynamic variability. \par
In this paper, we propose the \underline{D}istribution and \underline{R}elation \underline{A}daptive \underline{N}etwork (DRAN), a novel framework designed to dynamically adapt to temporal variations. DRAN facilitates more comprehensive representation learning and improved adaptability in spatio-temporal systems through a Spatial Factor Learner (SFL) for distribution adaptation and a Dynamic-Static Fusion Learner (DSFL) for adaptive modeling of static and dynamic spatial dependency.
The novelty and contributions of our method can be summarized as follows:\par
\begin{itemize}
\item
Temporal normalization can distort inter-node relations when directly applied to spatio-temporal data. To address this issue, we introduce a Spatial Factor Learner (SFL), which learns spatial scaling and shifting factors to restore the spatial structures affected by temporal normalization. In contrast to existing normalization methods that mainly adapt temporal statistics, SFL explicitly considers spatial relational consistency, enabling distribution adaptation without interfering with subsequent spatial dependency modeling.
\item
We propose a Dynamic-Static Fusion Learner (DSFL) that decomposes spatial representations into static and dynamic components and adaptively fuses them according to input variability. Existing decomposition-based methods typically combine different components through summation, concatenation, or globally shared fusion weights. In contrast, DSFL regulates the relative contributions of static and dynamic representations according to current spatio-temporal variability, allowing the model to emphasize dynamic relations during rapidly varying periods while retaining persistent spatial structures under relatively stable conditions.
\item 
Extensive experiments on real-world weather and traffic datasets demonstrate the effectiveness and generality of DRAN. Quantitative and qualitative analyses show that SFL better preserves spatial relational structures under different normalization strategies, while DSFL captures complementary static and dynamic dependencies and adjusts their contributions according to temporal variability.
\end{itemize}
\section{\label{sec:r}Related Work}
\subsection{Spatio-temporal Forecasting}
Spatio-temporal forecasting is a longstanding and important research task. Early methods such as Historical Average (HA) and ARIMA \cite{HA,ARIMA2} are interpretable but limited in capturing complex dependencies. With the advancement of deep learning, numerous frameworks have been proposed to enhance representation learning for spatio-temporal systems. \par
\textbf{1) Spatio-temporal dependency modeling}\par
\textit{Temporal dependency modeling}: For temporal dynamics, RNNs \cite{rnn1,dualts} perform well in short-term prediction, while transformers \cite{PatchTST,HMformer} and state-space models \cite{dstmamba} achieve strong results in both short- and long-term forecasting.\par
\textit{Spatial dependency modeling}: Spatial relations are commonly modeled through explicit graph structures \cite{seo2018structured}, learned similarity matrices \cite{liu2023spatio, Meta_Graph}, or causal-intervention-enhanced adaptive graphs~\cite{causalllm}. Static approaches typically learn adjacency matrices or node embeddings shared across samples from the training data \cite{bai2020adaptive, ijcai2022p328}. In contrast, dynamic approaches construct input-dependent graphs from sample-specific embeddings \cite{MixGT,lee2024testam}, while recent work further generates time-step-specific graphs using meta-parameter-enhanced node representations \cite{zou2026metadg}. With mechanisms such as graph attention or dynamic graph convolution, spatial relations can be inferred directly from node similarities \cite{zhang2024caformer} or generated through meta-learning strategies \cite{Meta_Graph}.\par
\textbf{2) Spatio-temporal architectures design}\par
Beyond individual modules, overall framework design also plays a critical role. 
Some methods organize temporal and spatial learners sequentially~\cite{liu2023spatio}, while others integrate temporal and spatial dependencies through joint architectures and cross-temporal spatial dynamic graph modeling ~\cite{Meta_Graph, dygtcyb}.
To improve interpretability and representation disentanglement, decomposition-based frameworks have gained increasing attention. Temporal decomposition methods separate time series into trend and seasonal components \cite{autoformer,dstmamba,stdn}. For example, STDN \cite{stdn} separates trends from seasonalities and processes them via parallel GRU blocks, while FWBNet \cite{FWBNet} leverages adaptive frequency–wavelet bases fused by cross-attention mechanisms. 
Spatial decomposition methods are explored across spatial scales \cite{fang2023learning}, spectral frequency \cite{decomp_spec,decomp_freq}, high- and low-frequency disentangled spatial representations \cite{qin2026adafre}, information redundancy \cite{redynet} or relational patterns \cite{decomp_direct,LEISN}. LEISN \cite{LEISN} models explicit and implicit relations through separate branches, whereas RGSL \cite{ijcai2022p328} jointly learns static and dynamic graph structures. However, existing methods primarily improve the learning of individual components, while the adaptive fusion of decomposed representations remains underexplored. Most approaches employ static fusion strategies, such as summation or concatenation, or utilize globally learnable weights \cite{li2023dynamic}, although the relative
importance of static and dynamic relations may vary with the input. This motivates an adaptive fusion mechanism that adjusts their contributions according to current spatio-temporal variations.\par
In this work, we propose a spatio-temporal forecasting framework that learns temporal dependencies under distribution shifts, decomposes spatial representations into static and dynamic components, and adaptively fuses them according to input variations.\par
\subsection{Adaptation to Distribution Shifts}
Normalization and de-normalization methods adapt to distribution shifts in time series by adjusting statistical properties such as mean and variance. RevIN \cite{kim2021reversible} introduces a learnable affine transformation to align the means and standard deviations of inputs with those of outputs. Non-stationary Transformer \cite{liu2022non} argues that existing stationary methods remove excessive information from time series, which hampers the model's ability to learn temporal dependencies. To address this, it introduces De-stationary Attention modules that aim to balance this trade-off. DAIN \cite{passalis2019deep} learns adaptive scaling and shifting factors through linear and gated layers. ST-Norm \cite{stnorm} proposes temporal and spatial normalization modules to refine high-frequency and local components. Dish-TS \cite{fan2023dish} performs separate normalization and de-normalization on lookback and horizon windows with learned means and standard deviations. 
Other methods adapt model components or representations under distribution shifts, including graph-relation updating under concept drift \cite{conti_tcyb}, multistream drift adaptation through data sharing \cite{multistream_tcyb}, dynamic filtering with closed-loop optimization \cite{pidada}, distribution-aware affine adaptation \cite{apt}, and test-time model adaptation \cite{tafas}.\par
Different from existing approaches that mainly adapt temporal statistics or model parameters, SFL is tailored to widely used temporal normalization schemes and explicitly preserves spatial-relation consistency among nodes.\par
\begin{figure*}[!t]
\centering
\includegraphics[width=0.72\textwidth]{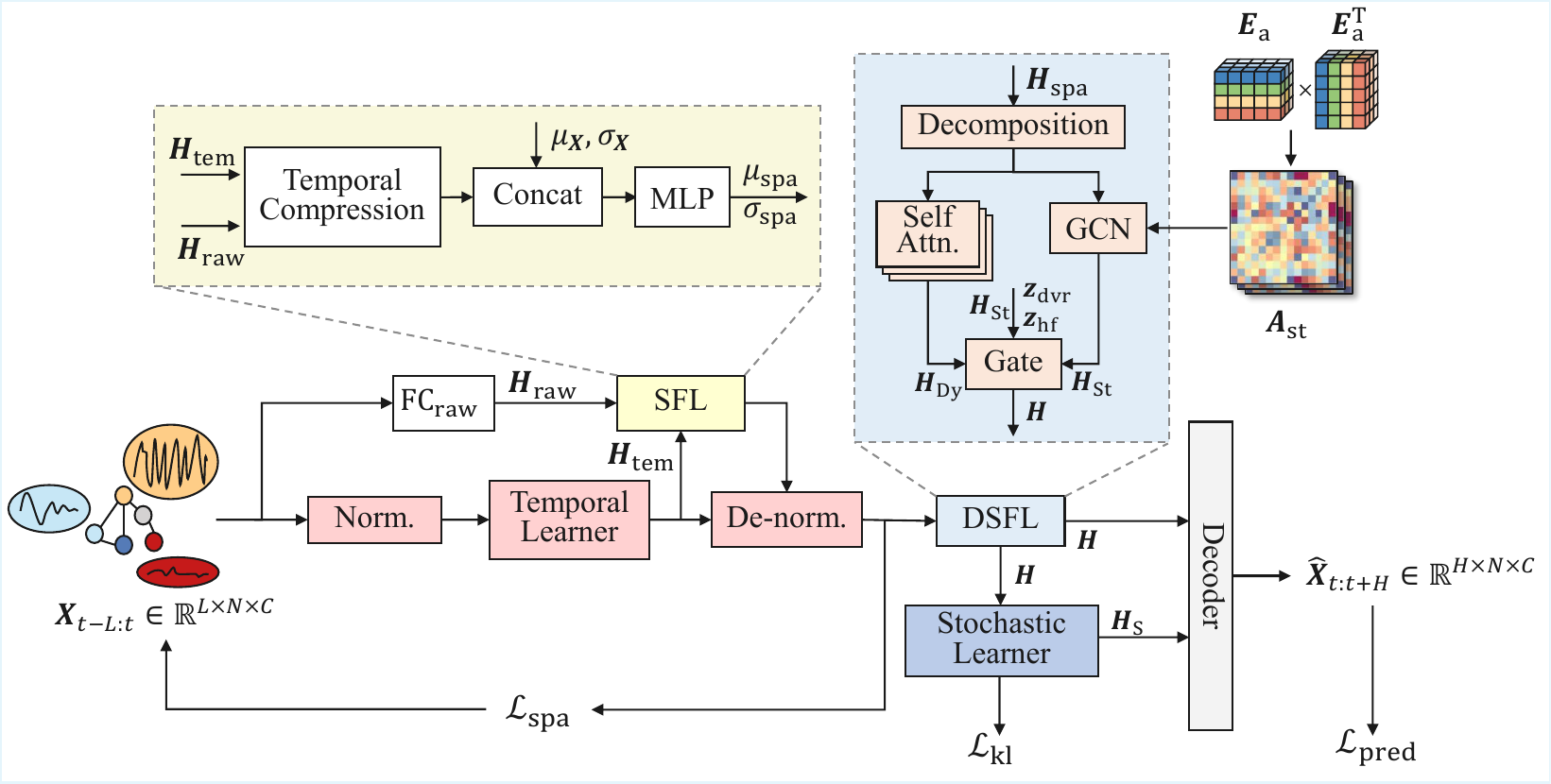}
\caption{\label{framework}
The overall architecture of DRAN. 
DRAN follows a temporal-spatial learning paradigm. 
SFL facilitates the normalization and de-normalization processes, denoted as ``Norm.'' and ``De-norm.'', respectively. 
DSFL decomposes the representations into dynamic and static components, and models spatial dependencies through attention mechanisms together with the adaptive adjacency matrix ${\bm A}_{\rm St}$ constructed from learnable node embeddings.}
\end{figure*}
\section{\label{sec:2}Methodology}
\subsection{Problem Definition and Preliminaries}
\begin{table}[htbp]
\caption{Summary of main notation and hyperparameters.}
\label{tab:notation}
\centering
\footnotesize
\setlength{\tabcolsep}{3pt}
\renewcommand{\arraystretch}{0.95}
\begin{tabular}{@{}cl@{}}
\hline
\multicolumn{1}{c}{\textbf{Symbol}} &
\multicolumn{1}{c}{\textbf{Description}} \\
\hline
$L,~H$ & Lookback and forecasting length. \\
$N,~C$ & Number of nodes and input variables. \\
$C'$ & Latent feature dimension. \\

$\bm{X}$, $\hat{\bm{X}}$ & Observed and predicted spatio-temporal sequence. \\
$\bm{\mu}_{X},\bm{\sigma}_{X}$ & Temporal mean and standard deviation. \\

$\bm{H}_{\rm raw}$ & Features extracted from unnormalized inputs. \\
$\bm{H}_{\rm tem}$ & Temporally encoded features. \\
$\bm{\mu}_{\rm spa},\bm{\sigma}_{\rm spa}$ & Spatial shift and scaling factors. \\
$\bm{H}_{\rm spa}$ & Spatially consistent features produced by SFL. \\

$\bm{E}_{\rm a}$ & Learnable adaptive node embeddings. \\
$\bm{A}_{\rm St}$ & Learned static adjacency matrix. \\
$\widetilde{\bm{A}}_{\rm St}$ & Normalized static graph operator. \\

$\bm{z}_{\rm St},\bm{z}_{\rm Dy}$ & Decomposed static and dynamic features. \\
$\bm{H}_{\rm St},\bm{H}_{\rm Dy}$ & Static and dynamic spatial representations. \\
$\gamma_l$ & Weight of the $l$-th attention layer. \\

$\bm{z}_{\rm dvr}$ & Variance ratio of first differences. \\
$\bm{z}_{\rm hf}$ & High-frequency energy ratio. \\
$\bm{z}_{\rm control}$ & Dynamic--static fusion gate. \\
$\bm{H}$ & Fused spatio-temporal representation. \\

$\bm{\mu}_{\rm sto},\bm{\sigma}_{\rm sto}$ & Parameters of the stochastic latent distribution. \\
$\bm{z}_{\rm l}$ & Sampled stochastic latent variable. \\
$\bm{H}_{\rm S}$ & Reconstructed stochastic representation. \\

$\alpha,\beta$ & Weights of $\mathcal{L}_{\rm spa}$ and $\mathcal{L}_{\rm kl}$. \\
$k$ & Number of dynamic self-attention layers. \\
\hline
\end{tabular}
\end{table}
\textbf{Spatio-temporal forecasting.} The observed variables of node $i$ at time step $t$ is represented by a $C$-dimensional vector $\bm{X}_{t,i} \in \mathbb{R}^{C}$. 
Each observation is the result of a stochastic process, drawn from a conditional distribution $\bm{X}_{t,i} \sim p_{t,i}(\bm{X}_{t,i}|\bm{X}_{t-1},\bm{X}_{t-2},\cdots, \bm{X}_{t-k},\cdots)$, where $\bm{X}_{t-k} \in \mathbb{R}^{N \times C}$ denotes the observations from $N$ nodes at time step $t-k$ in the spatio-temporal framework. 
For node $i$, $\bm{X}_{t,i}\in\mathbb{R}^{C}$ follows a
time-varying conditional distribution
$p_{t,i}(\bm{X}_{t,i}\mid\bm{X}_{<t})$. A distribution shift
occurs when
$D(p_{t_1,i},p_{t_2,i})>\delta$, where $D(\cdot, \cdot)$ denotes a distance function measuring the discrepancy between distributions, \(t_1\) and \(t_2\) denote two different time steps, and $\delta$ is a tolerance parameter that can be tuned according to the loss function and the expected drift magnitude. Under smooth drift, the
discrepancy may accumulate over longer intervals. \par
Given $\bm{X}_{t-L:t}\in\mathbb{R}^{L\times N\times C}$, the spatio-temporal forecasting task is
to learn a model $\rm F$ that predicts
$\bm{X}_{t:t+H}\in\mathbb{R}^{H\times N\times C}$.
Due to the time variance in spatio-temporal data generation, learning a model $\mathrm{F}$ that effectively handles shifting distributions is challenging. \par
\begin{algorithm}
\footnotesize
{\footnotesize\caption{DRAN Framework}\label{alg1}}
\begin{algorithmic}[1]
\REQUIRE Spatio-temporal dataset $Ds$, hyperparameter $\alpha$, $\beta$, lookback length $L$, horizon length $H$, max epoch
\REQUIRE model parameters $\theta$ of DRAN
\STATE Initialize model parameters $\theta$ and node embedding $E_{\rm a}$
\FOR {epoch = 1 to max epoch}
    \FOR{each batch $\textbf{\textit{X}}_{t-L:t},\textbf{\textit{X}}_{t:t+H} \in Ds$}
        \STATE Compute $\mu_{\bm{X}}$, $\sigma_{\bm{X}}$, $\bm{z}_{\rm dvr}$, $\bm{z}_{\rm hf}$
        \STATE $\bm{H}_{\rm raw}\xleftarrow{}\rm{FC_{raw}}(\bm{X}$)
        \STATE \textbf{Distribution adaptation:}
        \STATE Temporal normalization on $\bm{X}$ via Eq. (\ref{tnorm})
        \STATE $\bm{H}_{\rm tem}\xleftarrow{}\rm{Temporal\ learner}(\bm{\textit{X}},\mu_{\bm{X}},\sigma_{\bm{X}})$ 
        \STATE $\mu_{\rm spa}$, $\sigma_{\rm spa}\xleftarrow{}\rm{SFL}({\bm H}_{\rm raw},{\bm H}_{\rm tem}, \mu_{\bm{X}}, \sigma_{\bm{X}})$
        \STATE De-normalization: $\bm{H}_{\rm spa}\leftarrow\sigma_{\rm spa}\bm{H}_{\rm tem}+\mu_{\rm spa}$ 
        \STATE \textbf{DSFL:}
        \STATE $\bm{z}_{\rm St}\leftarrow{{\rm F}_{\rm GCN1}(\bm{\widetilde{A}}_{\rm St},\bm{H}_{\rm spa}})$, $\bm{z}_{\rm Dy}\leftarrow{\bm{H}_{\rm spa}-\bm{z}_{\rm St}}$
        \STATE $\bm{H}_{\rm St}\leftarrow{{\rm F}_{\rm GCN2}(\bm{\widetilde{A}}_{\rm St},\bm{z}_{\rm St}})$, $\bm{H}_{\rm Dy}\leftarrow\sum_{l=1}^k\gamma_l{\rm Self Attn}(\bm{z}_{\rm Dy})$
        \STATE $\bm{z}_{\rm control}\leftarrow{\rm Sigmoid}({\rm FC_g}({[\bm{H}_{\rm St},\bm{z}_{\rm dvr},\bm{z}_{\rm hf}]}))$
        \STATE $\bm{H}\leftarrow\bm{z}_{\rm control}\odot\bm{H}_{\rm Dy}+(\mathbf{1}-\bm{z}_{\rm control})\odot\bm{H}_{\rm St}$
        \STATE \textbf{Stochastic learner:}
        $\bm{H}_{\rm S}\xleftarrow{}{\rm Stochastic Learner}(\bm{H})$ 
        \STATE Compute $\mathcal{L}$ via Eq. (\ref{loss}) and optimize $\theta$
    \ENDFOR
\ENDFOR
\RETURN model parameters $\theta$
\end{algorithmic}
\end{algorithm}\textbf{Spatial message propagation.}
Most spatial learners update node $i$ by aggregating neighboring
representations:
\begin{equation}
    \bm{H}^{(l)}_i
    =
    \bm{H}^{(l-1)}_i
    +
    \sum_{j\in\mathcal{N}(i)}
    a^{(l)}_{ij}\bm{H}^{(l-1)}_j, \label{gnn}
\end{equation}
where $a^{(l)}_{ij}$ may be graph-based or attention-based.
Thus, spatial propagation depends on both node representations
and their pairwise relations.\par
\textbf{Series-level normalization.} Given a node-wise multivariate time series 
$\bm{X}_i \in \mathbb{R}^{L \times C}$, standard normalization rescales each series using its temporal mean $\mu_i$ and standard deviation $\sigma_i$:
\begin{equation}
    \bm{X}'_{i} = \frac{\bm{X}_i - \mu_i}{\sigma_i} \label{tnorm}.
\end{equation}
Although series-level normalization stabilizes temporal statistics, it inevitably rescales node features in a dimension-wise manner, thereby distorting the message propagation process defined in Eq. (\ref{gnn}). Therefore, it is necessary to design modules that transform the temporal representations learned from normalized features $H_{\mathrm{tem}}$ into spatially consistent representations $H_{\mathrm{spa}}$, so that spatial interactions can be modeled without distortion from node-wise rescaling.
The main notation and hyperparameters used throughout the paper are summarized in Table~\ref{tab:notation}.
The overall workflow of DRAN and the structure of its modules are depicted in Fig. \ref{framework} and detailed in Algorithm \ref{alg1}.\par
\subsection{\label{dis_ada}Temporal Dependency Modeling under Distribution Shifts}
DRAN adopts a temporal-first modeling strategy as distribution shifts are primarily reflected in the temporal statistics of individual nodes, whereas spatial dependency modeling relies on the relative representations across nodes. Accordingly, we first normalize the input series to mitigate temporal distribution shifts and then learn temporal dependencies from the normalized features, yielding $\bm{H}_{\rm tem}$. Before spatial propagation, SFL learns node-wise spatial factors to alleviate the distortion of inter-node relations introduced by temporal normalization and transforms $\bm{H}_{\rm tem}$ into spatially consistent representations $\bm{H}_{\rm spa}$.
Specifically, for each node, we normalize the input time series $\bm{X}$ by computing the temporal mean $\mu_{\bm{X}} \in \mathbb{R}^{1 \times N \times C}$ and standard deviation $\sigma_{\bm{X}} \in \mathbb{R}^{1 \times N \times C}$ over the window $[t-L, t)$, as defined in Eq.~(\ref{tnorm}).
To prevent the over-smoothing of temporal dynamics due to normalization, we adopt the de-stationary attention mechanism from the Non-stationary Transformer \cite{liu2022non} to obtain $\bm{H}_{\rm tem}$, which performs adaptive detrending to preserve non-stationary signals. \par
To preserve spatial consistency before feeding the features into the spatial layers, we de-normalize the node features $\bm{H}_{\rm tem}$ with our proposed SFL structure. SFL acts as a statistic factor generator $\Phi({\bm H}_{\rm raw}, {\bm H}_{\rm tem}, \mu_{\bm X}, \sigma_{\bm X} )$ to generate the spatial factors $\mu_{\rm spa}$ and $\sigma_{\rm spa}$, where $\bm{H}_{\rm raw}$ is obtained by applying a linear layer $\rm{FC_{raw}}$ to map the $C$ input variables to the $C'$-dimensional model space. In SFL, temporal distribution patterns are first extracted by applying a 1D convolution-based temporal compression to ${\bm H}_{\mathrm{tem}}$ and ${\bm H}_{\mathrm{raw}}$. Subsequently, spatial factors are generated based on ${\bm H}_{\mathrm{raw}}$, ${\bm H}_{\mathrm{tem}}$, $\mu_{\bm X}$, and $\sigma_{\bm X}$.
Finally, spatial factors $\mu_{\rm spa}$ and $\sigma_{\rm spa}$ are applied to de-normalize $\bm{H}_{\rm tem}$.
\begin{equation}
    \bm{H}_{\rm spa} = \sigma_{\rm spa}\bm{H}_{\rm tem}+\mu_{\rm spa},\label{n2}
\end{equation}
where $\bm{H}_{\rm spa}$ represents the de-normalized result of $\bm{H}_{\rm tem}$. \par
To mitigate the spatial-relation distortion introduced by normalization, we impose an additional reconstruction constraint that aligns $\bm{H}_{\rm spa}$ with  $\bm{H}_{\rm raw}$: 
\begin{equation}
\mathcal{L}_{\rm spa}
=
\mathbb{E}_{t,i}
\left[
\left\|
\bm{H}_{{\rm spa},t,i,:}
-
\bm{H}_{{\rm raw},t,i,:}
\right\|^2_2
\right].
\label{lspa}
\end{equation}
Let $\bm{H}_{\rm spa}^{*}$ denote a minimizer of this
reconstruction objective. As derived in Appendix~\ref{theory}, it satisfies:
\begin{equation}
\mathbb{E}_{t,i}
\left[
\left\|
\bm{H}^{*}_{{\rm spa},t,i,:}
-
\bm{H}_{{\rm raw},t,i,:}
\right\|_2^2
\right]
\leq
\mathbb{E}_{t,i}
\left[
\left\|
\bm{H}_{{\rm tem},t,i,:}
-
\bm{H}_{{\rm raw},t,i,:}
\right\|_2^2
\right].
\label{eq:node_reconstruction}
\end{equation}
Furthermore, the node-wise reconstruction error provides an upper bound on pairwise distance distortion. Therefore, $\mathcal{L}_{\rm spa}$ encourages SFL to preserve the spatial relations distorted by temporal normalization.
\subsection{DSFL for Spatial Learning}
After obtaining the spatially restored representation $\bm{H}_{\rm spa}$, DSFL decomposes it into a smooth static
component and a complementary residual that captures dynamic spatial variations, and adaptively fuses their spatial representations according to input variability. 
Specifically, a GCN layer $\rm{F}_{\rm GCN1}$ is employed as a low-pass filter in the graph spectral domain \cite{nt2019revisiting} for static-dynamic decomposition. The static adjacency matrix $\bm{A}_{\rm St} \in \mathbb{R}^{L\times N\times N}$ is derived from the adaptive node embedding $\bm{E}_{\rm a}$ to generate sample-shared and task-specific relations.
\begin{equation}
\begin{aligned}    
\bm{A}_{\rm St}&=\bm{E}_{\rm a}\bm{E}_{\rm a}^{\rm T},\\
\bm{\widetilde{A}_{\rm St}}&= {\bm D}^{-\frac{1}{2}}(\bm{A}_{\rm St}+I){\bm D}^{-\frac{1}{2}}\label{ast}\\
\bm{z}_{\rm St} &={\rm F}_{\rm GCN1}({\bm{\widetilde{A}_{\rm St}}}, \bm{H}_{\rm spa})\\
\bm{z}_{\rm Dy} &= \bm{H}_{\rm spa}-\bm{z}_{\rm St},
\end{aligned}
\end{equation}
where $\bm D$ denotes the degree matrix, $I$ is the identity matrix, and $\bm{\widetilde{A}_{\rm St}}$ is the normalized matrix derived from $\bm{A}_{\rm St}$. 
The static component $\bm{z}_{\rm St}$ is obtained through graph low-pass filtering and is expected to capture persistent spatial structures shared across samples. The dynamic component $\bm{z}_{\rm Dy}$ is explicitly defined as the residual part after removing $\bm{z}_{\rm St}$ from $\bm{H}_{\rm spa}$, and therefore represents high-frequency or sample-specific spatial variations that cannot be explained by the static component. The residual construction makes the two components complementary, satisfying
$\bm{z}_{\rm St}+\bm{z}_{\rm Dy}=\bm{H}_{\rm spa}$,
and explicitly separates persistent spatial structures from input-dependent variations.
\par
We employ branch-specific spatial learners according to the properties of the two components. Since $\bm{z}_{\rm St}$ is obtained through graph low-pass filtering and represents smooth
and persistent spatial patterns, it is processed by a GCN layer $\mathrm{F}_{\rm GCN2}$ with the sample-shared adjacency matrix $\bm{\widetilde{A}}_{\rm St}$ to obtain the static representation $\bm{H}_{\rm St}$. In contrast, self-attention is applied to the
residual dynamic component $\bm{z}_{\rm Dy}$ to infer
sample-specific node interactions. The outputs of the $k$ attention layers are then aggregated using learnable weights $\gamma_l$, integrating dynamic representations learned at different depths:
\begin{align}    
\bm{H}_{\rm St}& = {\rm F}_{\rm GCN2}({\bm{\widetilde{A}_{\rm St}}}, {\bm z}_{\rm St}),\\
\bm{H}_{\rm{Dy}}^{(l)}&={\rm Softmax}(\frac{\bm{z}_{\rm Dy}W_{\rm Q}^{(l)}(\bm{z}_{\rm Dy}W_{\rm K}^{(l)})^{T}}{\sqrt{C'}})\bm{z}_{\rm Dy}W_{\rm V}^{(l)}, \label{Dy} \\
\bm{H}_{\rm Dy}&=\sum_{l=1}^{k}\gamma_l\bm{H}_{\rm{Dy}}^{(l)} ,
\end{align}
where $W_{\rm K}^{(l)}$, $W_{\rm Q}^{(l)}$ and $W_{\rm V}^{(l)}$ are learnable projection matrices, and $\sqrt{C'}$ is the scaling factor used to stabilize gradients. 
To enable adaptive fusion between static and dynamic representations, we introduce a gating mechanism conditioned on static features and input-dependent dynamic variability metrics. The variance ratio of first differences $\bm{z}_{\rm dvr}$ and high-frequency energy ratio $\bm{z}_{\rm hf}$ are computed and concatenated with the static features to generate the control feature $\bm{z}_{\rm control}$ for gating.
\begin{equation}
    \begin{aligned}
    \bm{z}_{\rm control}&={\rm Sigmoid}({\rm FC_g}({[{\bm H}_{\rm St},\bm{z}_{\rm dvr},\bm{z}_{\rm hf}]})),\\
    \bm{H}&=\bm{z}_{\rm control}\odot\bm{H}_{\rm Dy}+(\mathbf{1}-\bm{z}_{\rm control})\odot\bm{H}_{\rm St}\label{eq1},
\end{aligned}
\end{equation}
where Sigmoid is the activation function, and $\rm{FC_{\rm g}}$ denotes the fully connected layer. Finally, we apply Eq. (\ref{eq1}) to fuse the dynamic and static features, yielding the final spatio-temporal representation $\bm{H}$. \par
\subsection{Forecasting and Optimization}
After obtaining the encoded spatio-temporal representation $\bm H$, we utilize a VAE to generate the stochastic components of the predictions. 
The features $\bm{H}$ are fed into the latent layers $\text{F}_{\text{lat}}(\cdot)$ to obtain the mean $\mu_{\rm sto}$ and standard deviation $\sigma_{\rm sto}$ of $\bm{H}$. Then, we sample the latent features $\bm{z}_{\rm l}$ from the distribution $\mathcal{N}(\mu_{\rm sto}, \sigma_{\rm sto}^2)$. These latent features $\bm{z}_{\rm l}$ are then processed through the reconstruction layer $\text{F}_{\text{rec}}(\cdot)$ to map back to the stochastic components of the time series. The processes are detailed below:
\begin{eqnarray}
    \mu_{\mathrm{sto}}, \sigma_{\mathrm{sto}} &=& {\mathrm F}_{\mathrm{lat}}(\bm{H}),\ 
    \bm{z}_{\mathrm{l}} \sim \mathcal{N}(\mu_{\mathrm{sto}},\sigma_{\mathrm{sto}}^2), \\
    \bm{H}_{\mathrm{S}} &=& {\mathrm F}_{\mathrm{rec}}(\bm{z}_{\mathrm{l}}),
\end{eqnarray}
where $\bm{H}_{\rm{S}}$ refers to the stochastic components of the forecast.
To ensure that the Stochastic Learner can effectively capture the stochastic components related to input dynamics, a constraint loss $\mathcal{L}_{\mathrm{kl}}$ based on the Kullback-Leibler (KL) divergence is imposed on the learned latent feature distribution to make it close to the standard normal distribution: \par
\begin{equation}
\mathcal{L}_{\rm kl}=-\frac{1}{2}\sum\left(1+\log \sigma_{\rm sto}^2-\mu_{\rm sto}^2-\sigma_{\rm sto}^2\right).
\end{equation}
The loss function $\mathcal{L}$ consists of three components: prediction error $\mathcal{L}_{\mathrm{pred}}$, spatial similarity loss $\mathcal{L}_{\mathrm{spa}}$, and the distribution loss $\mathcal{L}_{\mathrm{kl}}$. 
\begin{equation}
    \mathcal{L}=\mathcal{L}_{\rm pred}(\bm{X}_{t:t+H}, \hat{\bm{X}}_{t:t+H})+\alpha\mathcal{L}_{\rm spa}+\beta\mathcal{L}_{\rm kl} \label{loss},
\end{equation}
where $\mathcal{L}_{\mathrm{pred}}$ uses the mean absolute prediction error, $\alpha$ and $\beta$ are hyperparameters that balance the three loss terms, and $\hat{\bm{X}}$ represents the time series generated by the neural network. \par

Then $\bm{H}$ and $\bm{H}_{\mathrm{S}}$ are fed into the decoder to map the features to the forecasting horizons. In this paper, the decoder comprises stacks of FC layers.
\begin{equation}
    \hat{\bm{X}}_{t:t+H}={\rm Decoder}({\rm Concatenate}[\bm{H},\ \bm{H}_{\rm S}]).
\end{equation} \par
\subsection{Complexity Analysis}
SFL introduces temporal compression of $O(LNC')$ and the node-wise MLP of $O(NC'^2)$, giving an overall complexity of $O(LNC'+NC'^2)$, which is linear in both $L$ and $N$ for fixed $C'$. A standard $k$-layer spatial self-attention module requires $O(kL(N^2C'+NC'^2))$. Since the static adjacency matrix $\bm{A}_{\rm St}$ can be precomputed after training, the online complexity of DSFL, including two graph convolutional layers, dynamic spatial attention, and gating, is $O((k+2)LN^2C'+(k+3)LNC'^2)$. When $N>C'$, the dominant term is $O((k+2)LN^2C')$. Thus, SFL scales linearly with the number of nodes, while DSFL has the same order of standard spatial self-attention, with moderate constant-factor overhead.
\begin{table}[!t]
\setlength{\tabcolsep}{2pt}
\renewcommand{\arraystretch}{0.6} 
\centering
\caption{Balanced hyperparameter selection}
\footnotesize
\begin{tabular}{@{}c|c|c|c|c|c|c@{}}
\toprule
\makecell{Hyper\\-parameter}  & Weather & NYCBike1 & NYCBike2 & NYCTaxi & PeMS04 & PeMS08 \\ \midrule
$\alpha$ & 0.05     & 0.05      & 0.05      &1     &0.001    & 0.1    \\ 
$\beta$  & 0.5     & 0.5     & 0.5     & 0.5    & 5  & 0.05    \\ \bottomrule
\end{tabular}
\label{para_decide}
\end{table}
\begin{table}[t]
\centering
\caption{Dataset details}
\scriptsize
\renewcommand{\arraystretch}{0.6}
\begin{tabular}{c|cccr@{$\;\rightarrow\;$}l}
\toprule%
\textbf{Attributes}  & \textbf{\makecell{Duration time}}   & \textbf{Freq.} & \textbf{\makecell{Node\\number}}  & \multicolumn{2}{c}{\textbf{\makecell{Length\\(In$\rightarrow$Out)}}}  \\ \midrule
Weather  &01/01/2012$\sim$31/12/2022 &1 h              &263              &24&12            \\ \midrule
NYCBike1 &01/04/2014$\sim$30/09/2014  & 30 min         & 128                &19&1             \\
NYCBike2 &01/07/2016$\sim$29/08/2016  & 30 min           & 200                &35&1             \\
NYCTaxi  &01/01/2015$\sim$01/03/2015   & 30 min          & 200            &35&1             \\ \midrule
PeMS04   & 01/01/2018$\sim$28/02/2018  & 5 min            & 307              &12&12            \\
PeMS08   & 01/07/2016$\sim$31/08/2016  & 5 min          & 170              &12&12            \\ \bottomrule
\end{tabular}
\label{datasets}
\end{table}
\begin{table}[t]
\centering
\caption{Baseline methods.}
\footnotesize
\renewcommand{\arraystretch}{0.6}
\begin{tabular}{c|c|cc}
\toprule
\multicolumn{1}{c|}{Task type} & Methods  & Task adaptive & Dynamic adaptive \\
\midrule
\multirow{3}{*}{\makecell[c]{Time series\\forecasting}}
&DA-RNN \cite{dualts}  &\ding{55} & \ding{55}\\
&InfoTS \cite{infots}  & \ding{51} & \ding{51} \\
&AutoTCL \cite{parats}  & \ding{51} & \ding{51} \\
\midrule 
\multirow{15}{*}{\makecell[c]{Spatio-temporal\\forecasting}}
&TGCN\cite{zhao2019t}          & \ding{55}     & \ding{55} \\
&STGCN\cite{yu2018spatio}  & \ding{55} & \ding{55} \\
&GCGRU \cite{seo2018structured}   & \ding{55} & \ding{55} \\
&AGCRN\cite{bai2020adaptive}  & \ding{51} & \ding{55} \\
&ASTGCN\cite{guo2019attention}   & \ding{55} & \ding{51} \\
&DCRNN\cite{li2018diffusion} & \ding{55} & \ding{55} \\
&STAEformer\cite{liu2023spatio}   & \ding{51} & \ding{51} \\
&ST-SSL\cite{ji2023spatio}   & \ding{55} & \ding{51} \\
&MegaCRN \cite{Meta_Graph}  & \ding{51} & \ding{51} \\
&RGSL \cite{ijcai2022p328}   & \ding{51} & \ding{55} \\
&TESTAM \cite{lee2024testam}  & \ding{51} & \ding{51} \\
&MemDA \cite{memda}  & \ding{51} & \ding{51} \\
&DST-Mamba \cite{dstmamba} &\ding{51} &\ding{51} \\
&STDN \cite{stdn} &\ding{51} &\ding{51} \\
&Ours    & \ding{51} & \ding{51} \\
\bottomrule
\end{tabular}
\label{baseline}
\end{table}

\section{\label{sec:3}Experimental Setup}
In this section, we detail our experimental setup, including the training process, the hyperparameter settings for our networks, the datasets and the baselines.\par
\subsection{Training Details}
We train DRAN using the Adam optimizer with a learning rate of 0.001, batch size 32, and 100 training epochs. The hyperparameters $\alpha$ and $\beta$ in Eq. (\ref{loss}) are fine-tuned experimentally to account for the stochastic nature and uncertainties of the datasets (see Table \ref{para_decide}). The selection process is described in Appendix Table \ref{hyper_select}. All the experiments are conducted using five random seeds from the set $\{31, 32, 33, 34, 35\}$ to obtain the average performance and standard deviation. Detailed training configurations are listed in Appendix \ref{training}.
\subsection{Datasets}
We evaluate spatio-temporal forecasting performance on six real-world datasets covering weather and traffic systems: Weather, NYCBike1, NYCBike2, NYCTaxi, PeMS04, and PeMS08. The Weather dataset is derived from the ERA5 hourly temperature data \cite{era5}. The three NYC datasets contain bike and taxi flow records collected in New York City \cite{ji2023spatio}, while PeMS04 and PeMS08 consist of traffic flow measurements collected by loop detectors on California highways \cite{chen2001freeway}. The main dataset statistics are summarized in Table~\ref{datasets}, with further details provided in Appendix~\ref{dataset_des} and Table~\ref{dataset_add}.\par
\subsection{Baselines and Setup}
As shown in Table \ref{baseline}, we compare our method with several baselines, including state-of-the-art multivariate time series and spatio-temporal forecasting models. 
For fairness, all baselines are evaluated under the same data splits, input-output lengths, metrics, and training protocol. For each baseline, we use the official or publicly available implementation when available, and tune its key hyperparameters on the validation set for each dataset; the candidate ranges and final selected values are provided in Appendix \ref{baseline_app}.
Model performance is evaluated using Mean Absolute Error (MAE) and Mean Absolute Percentage Error (MAPE), and a distributional metric based on the Wasserstein Distance (WD), where MAE and MAPE measure point-wise errors and WD evaluates distributional consistency. To evaluate statistical reliability, we conduct two-sided paired $t$-tests between DRAN and the strongest competing baseline over five random seeds, with $^{\dagger}$ indicating $p<0.05$. Metric definitions are included in Appendix \ref{metrics}. \par
\begin{table*}[!t]
\centering
\caption{The prediction results on weather, NYCBike1 and NYCBike2 datasets.}
\renewcommand{\arraystretch}{0.8} 
\setlength{\tabcolsep}{4pt}
\footnotesize
\begin{tabular}{@{}c|
ccc|
ccc|
ccc@{}}
\toprule
\multirow{2}{*}{Model}
& \multicolumn{3}{c}{Weather}
& \multicolumn{3}{c}{NYCBike1}
& \multicolumn{3}{c}{NYCBike2} \\
\cmidrule(lr){2-4}\cmidrule(lr){5-7}\cmidrule(lr){8-10}
& MAE$\downarrow$  & MAPE(\%)$\downarrow$ &WD$\downarrow$  & MAE$\downarrow$  & MAPE(\%)$\downarrow$ &WD$\downarrow$  & MAE$\downarrow$  & MAPE(\%)$\downarrow$ &WD$\downarrow$ \\
\midrule
DA-RNN \cite{dualts}      &5.492\std{0.725}	&1.874\std{0.276}	&3.758\std{0.670}
&15.773\std{2.406}	&61.948\std{7.968}	&5.970\std{0.955}
&15.159\std{3.591}	&63.687\std{9.591}	&3.567\std{1.106}   \\
InfoTS \cite{infots} &1.274\std{0.137}	&0.435\std{0.053}	&0.492\std{0.035}   &6.526\std{0.336}	&33.681\std{1.831}	&\underline{0.923\std{0.044}}   &6.259\std{0.368}	&30.628\std{1.626}	&\underline{0.455\std{0.044}}  \\
AutoTCL \cite{parats} &1.194\std{0.022}	&0.408\std{0.008}	&\underline{0.420\std{0.019}} &6.213\std{0.213}	&28.824\std{0.808}	&0.978\std{0.024}	&5.772\std{0.246}	&28.639\std{0.993}	&0.476\std{0.026}	 \\ \midrule
STGCN \cite{yu2018spatio} &2.074\std{1.004}	&0.709\std{0.396}	&1.960\std{0.918}  &17.141\std{0.142}	&58.498\std{1.610}	&11.634\std{1.827}
&17.297\std{0.244}	&55.595\std{0.805}	&7.990\std{0.046}	  \\
TGCN \cite{zhao2019t} &1.864\std{0.797}	&0.635\std{0.311}	&1.769\std{0.704}   &7.544\std{0.311}	&34.848\std{1.287}	&3.166\std{0.713}   &11.488\std{8.575}	&36.789\std{7.720}	&3.402\std{3.165} \\
MemDA \cite{memda} &1.714\std{0.039}	&2.479\std{4.638}	&1.803\std{2.362}	  &6.994\std{0.359}	&29.792\std{1.024}	&1.059\std{0.097}	    &6.666\std{0.334}	&29.653\std{0.955}	&3.508\std{0.051}	 \\
ASTGCN \cite{guo2019attention} &1.517\std{0.122}	&0.519\std{0.047}	&1.471\std{0.127}   &6.712\std{0.371}	&31.590\std{1.102}	&1.607\std{0.257}	   &6.190\std{0.290}	&28.764\std{0.786}	&2.293\std{0.126}  \\
TESTAM \cite{lee2024testam} &1.481\std{1.203} & 0.507\std{0.415} &1.203\std{0.264}   &6.579\std{0.759} &30.786\std{2.749} &  1.863\std{0.962}  &6.365\std{0.866} &28.143\std{2.948} &  2.665\std{0.482}\\
DST-Mamba \cite{dstmamba} &0.886\std{0.095}	&0.302\std{0.037}	&0.548\std{0.018}   &6.252\std{3.162}	&24.537\std{5.125}	&3.470\std{0.765}	   &5.171\std{0.202}	&25.046\std{0.846}	&2.390\std{0.577} \\
AGCRN \cite{bai2020adaptive} &0.990\std{0.020}	&0.338\std{0.008}	&0.966\std{0.021}   &6.423\std{0.634}	&30.312\std{1.526}	&7.356\std{0.838}  &10.377\std{4.095}	&34.516\std{3.210}	&9.377\std{0.999} \\
GCGRU \cite{seo2018structured} &1.013\std{0.038}	&0.346\std{0.015}	&0.986\std{0.034}  &5.544\std{0.206}	&27.080\std{1.112}	&1.242\std{0.184}    &6.643\std{2.753}	&24.024\std{2.509}	&1.204\std{0.322} \\
DCRNN \cite{li2018diffusion} &0.979\std{0.022}	&0.334\std{0.008}	&0.955\std{0.021}  &7.673\std{0.326}	&33.218\std{1.528}	&1.556\std{0.221}   &8.517\std{1.929}	&34.611\std{6.209}	&1.303\std{0.258}  \\
STAEformer \cite{liu2023spatio} &3.728\std{2.117}	&1.284\std{0.819}	&3.683\std{2.110}   &8.105\std{5.580}	&32.309\std{12.484}	&3.858\std{4.536}	  &5.283\std{0.205}	&25.307\std{0.643}	&1.590\std{0.082}  \\
STDN \cite{stdn} &1.019\std{0.040}	&0.348\std{0.015}	&0.476\std{0.089}   &5.202\std{0.158}	&25.517\std{0.631}	&8.513\std{1.886}   &5.223\std{0.152}	&25.021\std{0.724}	&8.942\std{1.891}  \\
MegaCRN \cite{Meta_Graph} &0.947\std{0.024}	&0.323\std{0.010}	&0.924\std{0.023}	   &5.198\std{0.160}	&25.806\std{0.550}	&1.510\std{0.157}   &5.057\std{0.156}	&24.452\std{0.608}	&1.334\std{0.158} \\
RGSL \cite{ijcai2022p328}   &\underline{0.727\std{0.003}}	&\underline{0.248\std{0.001}}	&0.701\std{0.003}   &5.253\std{0.171}	&26.074\std{0.548}	&3.268\std{0.439}  &5.155\std{0.204}	&24.832\std{0.706}	&2.214\std{0.456}	 \\
ST-SSL \cite{ji2023spatio} &1.394\std{0.035}	&2.872\std{5.871}	&2.073\std{3.747}	  &\underline{5.115\std{0.158}}	&\underline{24.297\std{0.354}}	 &4.357\std{1.390}   &\underline{4.910\std{0.191}}	&\underline{22.025\std{0.923}}	& 2.292\std{1.172}  \\ \midrule
DRAN (ours)&{\textbf{0.676\std{0.005}}}	&{\textbf{0.224\std{0.002}}}
 &{\textbf{0.392\std{0.011}} } &{\textbf{5.046\std{0.141}}}&{\textbf{23.939\std{0.555}}}	&{\textbf{0.415\std{0.148}}}
 &{\textbf{4.845\std{0.203}}}	&{\textbf{21.948\std{0.629}}}	&{\textbf{0.437\std{0.039}}}
  \\
\scriptsize $p$-value vs. 2nd& \scriptsize 0.001$^{\dagger}$ & \scriptsize 0.001$^{\dagger}$ & \scriptsize 0.034$^{\dagger}$ &\scriptsize 0.001$^{\dagger}$ &\scriptsize 0.003$^{\dagger}$ &\scriptsize 0.001$^{\dagger}$ &\scriptsize 0.008$^{\dagger}$ &\scriptsize 0.001$^{\dagger}$ &\scriptsize 0.044$^{\dagger}$ \\
\bottomrule             
\end{tabular}
\begin{tablenotes} 
    \item Results in \textbf{bold} indicate the best performance, and \underline{underlined} results indicate the second best performance. The $p$-values are computed by two-sided paired $t$-tests between DRAN and the strongest competing baseline over random seeds. $^{\dagger}$ indicates statistical significance with $p<0.05$. Values smaller than 0.001 are reported as $<0.001$.
\end{tablenotes}
\label{n_r_1}
\end{table*}
\begin{table*}[!t]
\centering
\caption{The prediction results on NYCTaxi, Pems04 and Pems08 datasets.}
\renewcommand{\arraystretch}{0.9} 
\setlength{\tabcolsep}{2pt}  
\footnotesize
\begin{tabular}{@{}c|
ccc|
ccc|
ccc@{}}
\toprule
\multirow{2}{*}{Model}
& \multicolumn{3}{c}{NYCTaxi}
& \multicolumn{3}{c}{PeMS04}
& \multicolumn{3}{c}{PeMS08} \\
\cmidrule(lr){2-4}\cmidrule(lr){5-7}\cmidrule(lr){8-10}
& MAE$\downarrow$ &MAPE(\%)$\downarrow$ &WD$\downarrow$  & MAE$\downarrow$ &MAPE(\%)$\downarrow$ &WD$\downarrow$  & MAE$\downarrow$ &MAPE(\%)$\downarrow$ &WD$\downarrow$ \\
\midrule
DA-RNN \cite{dualts}      &26.682\std{8.176}	&69.028\std{12.085}	&11.912\std{9.342}
&138.741\std{19.896}	&192.640\std{37.494}	&80.935\std{12.140}
&109.276\std{13.097}	&123.719\std{33.187}	&69.984\std{11.873} \\
InfoTS \cite{infots} &13.286\std{1.695}	&21.505\std{0.562}	&\underline{1.670\std{0.038}}	 &25.801\std{0.408}	&20.058\std{1.193}	&11.188\std{0.644}  &23.604\std{1.243}	&14.642\std{0.583}	&12.469\std{1.371}	  \\
AutoTCL \cite{parats} &13.119\std{1.729}	&21.601\std{0.509}	&1.768\std{0.060}  &23.814\std{0.043}	&17.355\std{0.200}	&10.374\std{0.180}	   &20.879\std{0.088}	&13.006\std{0.121}	&10.074\std{0.094}  \\ \midrule
STGCN \cite{yu2018spatio} &25.227\std{4.222}	&29.533\std{5.694}	&8.563\std{10.891}  &29.114\std{5.555}	&23.466\std{7.692}	&26.608\std{4.168}   &27.173\std{10.017}	&16.585\std{7.589}	&13.060\std{11.726}  \\
TGCN \cite{zhao2019t} &23.227\std{10.239}	&44.356\std{10.694}	&5.563\std{6.891}
&34.853\std{0.263}	&28.078\std{0.743}	&30.670\std{0.337}
&37.969\std{0.324}	&30.431\std{0.690}	&23.658\std{0.523}  \\
MemDA \cite{memda} &20.852\std{0.848}	&32.591\std{3.476}	&4.165\std{0.715}  &20.037\std{0.150}	&\underline{11.969\std{0.100}}	&6.585\std{0.429}  &16.500\std{0.106}	&9.481\std{0.372}	&6.032\std{0.156}	 \\
ASTGCN \cite{guo2019attention} &16.702\std{2.461}	&22.832\std{4.118}	&3.231\std{1.722}  &23.565\std{0.843}	&16.900\std{0.890}	&21.326\std{1.004}
&20.308\std{0.837}	&27.280\std{24.754}	&9.957\std{1.891}  \\
TESTAM \cite{lee2024testam} &16.826\std{2.339} &25.846\std{3.815} &2.633\std{0.437}   &19.331\std{0.481} &12.098\std{0.419} &5.492\std{0.469}    &16.379\std{0.207} &9.342\std{0.243} &  4.841\std{0.563}\\
DST-Mamba \cite{dstmamba} &15.766\std{1.302}	&19.126\std{0.745}	&7.095\std{3.771}  &20.706\std{2.160}	&15.798\std{1.992}	&\underline{4.899\std{1.153}}	    &17.530\std{0.809}	&10.529\std{0.634}	&\underline{4.316\std{0.946}}	 \\
AGCRN \cite{bai2020adaptive} &20.928\std{4.316}	&30.583\std{9.173}	&11.281\std{5.273}   &19.205\std{0.140}	&12.937\std{0.035}	&16.327\std{0.152}   &16.848\std{0.666}	&9.890\std{0.545}	&6.259\std{0.419}  \\
GCGRU \cite{seo2018structured} &11.215\std{2.099}	&22.593\std{6.915}	&2.113\std{0.656}  &25.840\std{0.038}	&17.788\std{0.442}	&23.968\std{0.031}  &18.467\std{0.138}	&10.469\std{0.209}	&4.901\std{0.184}  \\
DCRNN \cite{li2018diffusion} &11.005\std{2.665}	&21.941\std{4.257}	&2.834\std{2.448}   &25.835\std{0.017}	&17.996\std{0.498}	&24.010\std{0.031}
&20.841\std{0.056}	&11.868\std{0.108}	&6.609\std{0.297}  \\
STAEformer \cite{liu2023spatio} &11.043\std{1.230}	&18.046\std{0.710}	&2.357\std{0.336}		&\underline{18.241\std{0.073}}	&12.064\std{0.072}	&5.332\std{0.104}   &\underline{13.538\std{0.034}}	& 8.857\std{0.017}	&4.625\std{0.042}	  \\
STDN \cite{stdn} &11.479\std{1.392}	&19.044\std{1.115}	&25.273\std{6.499}	   &20.920\std{0.361}	&14.379\std{1.494}	&12.273\std{0.911}    &18.088\std{0.713}	&10.951\std{0.791}	&8.511\std{1.101}  \\
MegaCRN \cite{Meta_Graph} &10.973\std{1.235}	&18.002\std{0.428}	&1.878\std{0.161}   &19.376\std{0.127}	&12.808\std{0.098}	&6.297\std{0.199}	    &15.597\std{0.213}	&\underline{8.835\std{0.096}}	&4.903\std{0.165} \\
RGSL \cite{ijcai2022p328} &11.948\std{1.327}	&19.327\std{0.367}	&2.207\std{0.217}  &19.587\std{0.073}	&11.930\std{0.150}	&6.910\std{0.160}
&16.209\std{0.126}	&9.247\std{0.096}	&5.556\std{0.131}  \\
ST-SSL \cite{ji2023spatio} &\underline{11.218\std{1.142}}	&\underline{17.159\std{0.574}}	&9.228\std{1.501}  &23.146\std{0.961}	&13.857\std{1.498}	&7.918\std{1.403}   &18.989\std{0.637}	&14.813\std{6.863}	&6.307\std{0.409}  \\ \midrule
DRAN (ours) &{\textbf{10.721\std{0.980}}}	&{\textbf{16.845\std{0.530}	}}&{\textbf{0.750\std{0.146}}}
&{\textbf{18.132\std{0.008}}}	&{\textbf{11.879\std{0.097}}	}&{\textbf{4.687\std{0.187}}	}
    &{\textbf{13.366\std{0.117}}}	&{\textbf{8.733\std{0.090}}	}&{\textbf{4.180\std{0.075}}}	  \\
$p-$value vs. 2nd &\scriptsize 0.001$^{\dagger}$ &\scriptsize 0.019$^{\dagger}$ &\scriptsize 0.001$^{\dagger}$ &\scriptsize 0.045$^{\dagger}$ &\scriptsize 0.036$^{\dagger}$ &\scriptsize 0.405 &\scriptsize 0.024$^{\dagger}$ &\scriptsize 0.004$^{\dagger}$ &\scriptsize 0.782 \\
\bottomrule             
\end{tabular}
\begin{tablenotes} 
    \item Results in \textbf{bold} indicate the best performance, and \underline{underlined} results indicate the second best performance. The $p$-values are computed by two-sided paired $t$-tests between DRAN and the strongest competing baseline over random seeds. $^{\dagger}$ indicates statistical significance with $p<0.05$. Values smaller than 0.001 are reported as $<0.001$.
\end{tablenotes}
\label{n_r_2}
\end{table*}
\begin{figure*}[!t]
\centering
\tiny
\includegraphics[width=\linewidth]{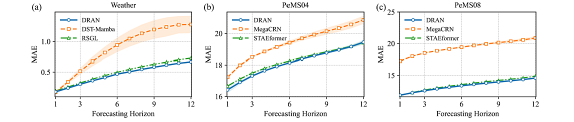}
\caption{\label{horizon}
Horizon-wise MAE comparison of DRAN and two strong baselines on Weather, PeMS04, and PeMS08. The curves and shaded regions denote the mean and standard deviation over five random seeds, respectively.
}
\end{figure*}
\begin{figure*}[htbp]
\centering
\includegraphics[width=0.72\textwidth]{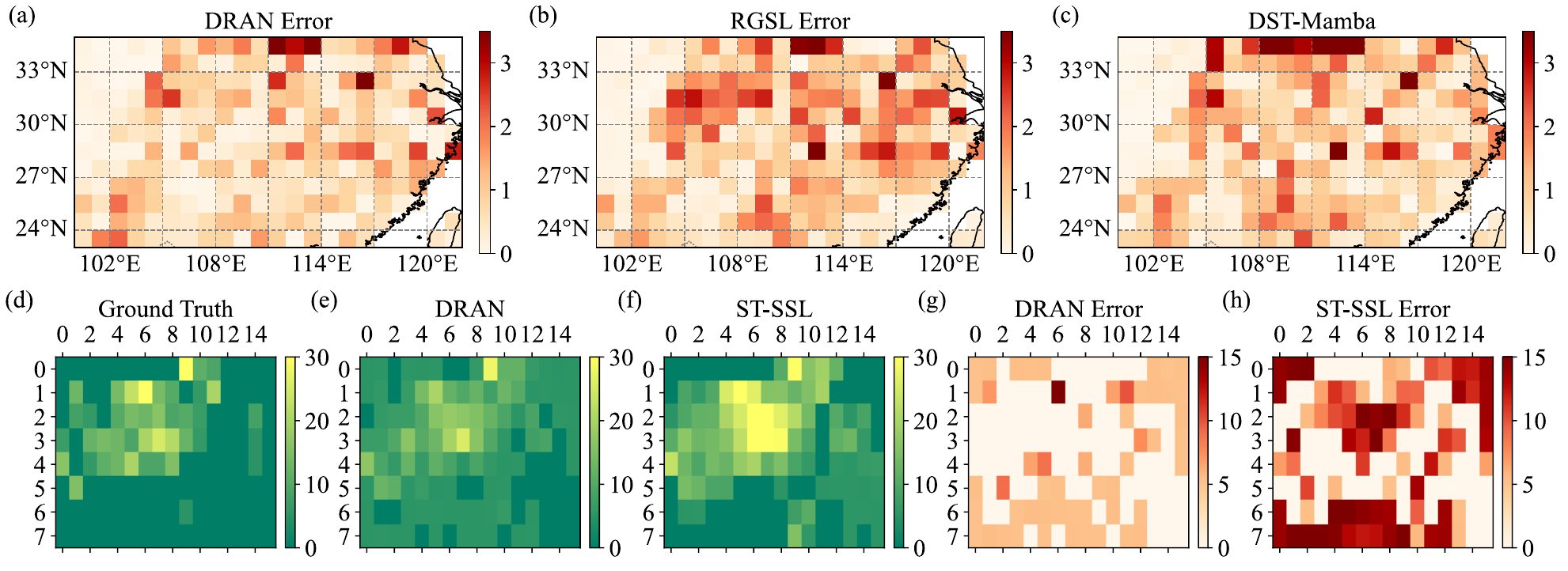}
\caption{\label{weather}Visualization of prediction results.
(a)--(c) Absolute prediction errors of DRAN, RGSL, and DST-Mamba
on the Weather dataset, respectively.
(d)--(h) Prediction results on NYCBike1, where (d) shows the ground truth,
(e) and (f) show the predictions of DRAN and ST-SSL, respectively,
and (g) and (h) present their corresponding absolute prediction errors.}
\end{figure*}
\begin{figure}[t]
\centering
\tiny
\includegraphics[width=\linewidth]{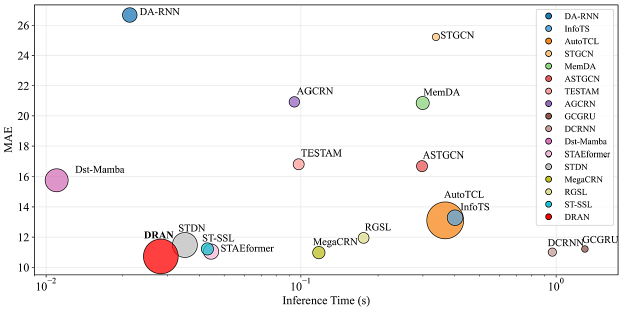}
\caption{\label{infer}Computational cost analysis on the NYCTaxi dataset. The x-axis denotes inference time (in seconds), and the y-axis represents MAE. Bubble size indicates the number of model parameters. 
DRAN achieves an accuracy-efficiency trade-off, obtaining lower MAE with competitive inference time compared to existing methods.}
\end{figure}
\begin{table}[!t]
\centering
\caption{Quantitative comparison of spatial-distribution preservation.}
\label{tab:spatial_preservation}
\footnotesize
\setlength{\tabcolsep}{3pt}
\renewcommand{\arraystretch}{0.95}
\begin{tabular}{c|cc|cc}
\toprule
\multirow{2}{*}{Dataset}
& \multicolumn{2}{c|}{$\bm{H}_{\rm tem}$ vs.\ $\bm{X}$}
& \multicolumn{2}{c}{$\bm{H}_{\rm spa}$ vs.\ $\bm{X}$} \\ \cmidrule(lr){2-5}
& PDD$\downarrow$ & ${\rm WD}_{\rm rel}$ $\downarrow$
& PDD$\downarrow$ & ${\rm WD}_{\rm rel}$ $\downarrow$ \\
\midrule
Weather
& 0.491\std{0.205}
& 0.177\std{0.056} 
& 0.180\std{0.054}
& 0.086\std{0.029} \\
NYCBike1
& 0.481\std{0.075}
& 0.470\std{0.087}
& 0.215\std{0.071}
& 0.129\std{0.078} \\

NYCBike2
& 0.177\std{0.034}
& 0.143\std{0.036}
& 0.168\std{0.071}
& 0.103\std{0.072} \\

NYCTaxi
& 0.130\std{0.024}
& 0.109\std{0.029}
& 0.118\std{0.040}
& 0.079\std{0.035} \\

PeMS04
& 0.198\std{0.020}
& 0.437\std{0.233}
& 0.171\std{0.202}
& 0.137\std{0.024} \\

PeMS08
& 0.131\std{0.018}
& 0.112\std{0.022}
& 0.110\std{0.030}
& 0.067\std{0.026} \\
\bottomrule
\end{tabular}
\begin{tablenotes}
    \item Pairwise Distance Distortion (PDD) and Wasserstein Distance (WD) are computed on spatial relation matrices of each time step. Lower values indicate better preservation of the original spatial structure.
\end{tablenotes}
\end{table}
\section{\label{sec:4}Experimental Results}
\begin{table*}[t]
\centering
\caption{Robustness comparison under test-time perturbations and naturally
high-dynamic periods.}
\label{tab:robustness}
\footnotesize
\setlength{\tabcolsep}{3.6pt}
\renewcommand{\arraystretch}{1.08}
\begin{tabular}{c|c|c|ccc|ccc|c}
\toprule
\multirow{2}{*}{Dataset} & \multirow{2}{*}{Model}
& \multirow{2}{*}{Clean}
& \multicolumn{3}{c|}{Gaussian Noise}
& \multicolumn{3}{c|}{Missing Sensors}
& High-dynamic \\
\cmidrule(lr){4-6}
\cmidrule(lr){7-10}
& &
& $\sigma=0.1$
& $\sigma=0.2$
& $\sigma=0.3$
& $\rho=10\%$
& $\rho=20\%$
& $\rho=30\%$
& Top 10\% \\
\midrule
\multirow{2}{*}{Weather}
& DRAN
& \textbf{0.676\std{0.005}}
& \textbf{1.369\std{0.060}}
& \textbf{2.039\std{0.037}}
& \textbf{2.557\std{0.031}}
&\textbf{1.227\std{0.032}}	&\textbf{1.328\std{0.062}}	&\textbf{1.397\std{0.051}}
& \textbf{0.948\std{0.009}} \\

& RGSL
& 0.727\std{0.003}
& 1.481\std{0.004}
& 2.161\std{0.006}
& 2.871\std{0.004}
&1.315\std{0.004}	&1.385\std{0.016}	&1.415\std{0.046}
& 0.973\std{0.003} \\
\midrule

\multirow{2}{*}{PeMS08}
& DRAN
& \textbf{13.366\std{0.117}} 
& \textbf{14.366\std{0.040}}
& \textbf{16.061\std{0.061}}
& \textbf{18.483\std{0.162}}
& \textbf{21.919\std{1.719}}
& \textbf{29.591\std{2.030}}
& \textbf{39.017\std{1.771}}
& \textbf{19.325\std{0.035}} \\

& STAEformer
& 13.538\std{0.034}
& 14.854\std{0.110}
& 17.168\std{0.442}
& 20.344\std{0.961}
& 22.412\std{0.707}
& 32.739\std{1.328}
& 41.851\std{1.805}
& 19.434\std{0.095} \\
\bottomrule
\end{tabular}
\begin{tablenotes}
    \item  Results are MAE over five random seeds. All models are trained on clean data and evaluated without retraining. Lower values are better.
\end{tablenotes}
\end{table*}
\begin{table*}[!t]
\centering 
\renewcommand{\arraystretch}{0.6} 
\caption{\label{dis-tab}The effectiveness of SFL on various temporal normalization methods.}
\footnotesize
\begin{tabular}{@{}c|cccccc@{}}
\toprule
\multirow{2}{*}{{Strategies}}
&Weather & {NYCBike1} & {NYCBike2} & {NYCTaxi} &{PeMS04} &{PeMS08}\\
 &MAE$\downarrow$ &MAE$\downarrow$ &MAE$\downarrow$ &MAE$\downarrow$ &MAE$\downarrow$ &MAE$\downarrow$          \\ \midrule
+RevIN                 &0.844\std{0.029}  &5.419\std{0.178}     &5.386\std{0.187}       &11.656\std{1.486}       &18.875\std{0.196}       &13.602\std{0.282}     \\ \midrule
+DAIN                  &1.004\std{0.453}     &5.541\std{0.189}     &5.424\std{0.174} &11.867\std{1.401}	    &18.695\std{0.231}       &13.704\std{0.070}       \\
+DAIN+SFL             &0.840\std{0.069}  &5.291\std{0.182}     &5.276\std{0.017}     &11.517\std{1.307}   &18.305\std{0.346}      &13.366\std{0.116}       \\ \midrule
+Non-st              &1.194\std{0.002}    &5.502\std{0.163}     &5.266\std{0.229}     &12.426\std{1.016}    &18.642\std{0.064}  &13.995\std{0.555}   \\
+Non-st+SFL         & 0.676\std{0.005}    &5.046\std{0.141}    &4.845\std{0.203}      & 10.721\std{0.980}   & 18.132\std{0.008}    & 13.366\std{0.117}        \\ \midrule
+Dish-TS               &1.850\std{0.001}    &5.491\std{0.182}        &5.564\std{0.193}         &11.701\std{1.276}    &18.743\std{0.186} &13.664\std{0.014}      \\
+Dish-TS+SFL          &0.813\std{0.011}  &5.268\std{0.182}    &5.341\std{0.177}   &11.303\std{1.296}         & 18.170\std{0.092}       &13.416\std{0.067} \\ \midrule
+ST-Norm  &0.868\std{0.038}   &5.393\std{0.163} &5.536\std{0.215} &11.643\std{1.366} & 18.317\std{0.306} & 13.680\std{0.120} \\
+ST-Norm+SFL  &0.834\std{0.023}   &5.307\std{0.066}  &5.439\std{0.217} &11.172\std{1.259}	&18.241\std{0.221}
&13.514\std{0.116}
\\ \bottomrule
\end{tabular}
\end{table*}

\subsection{Comparison Results}
\textbf{Numerical results:} As shown in Tables \ref{n_r_1} and \ref{n_r_2}, DRAN consistently outperforms the baselines, achieving the best overall results on both point error metrics and distributional evaluation metrics. This indicates that DRAN not only minimizes absolute prediction errors but also improves the distributional alignment between predictions and ground-truth values. 
The paired t-tests against the strongest competing baseline further confirm statistically significant improvements on most dataset–metric pairs. The nonsignificant WD improvements on PeMS04 and PeMS08 can be partly attributed to the large seed-wise fluctuations of the competing baselines, which increase the variance of the paired differences. \par
Fig.~\ref{horizon} presents the horizon-wise forecasting errors. DRAN achieves the lowest MAE at most steps and remains competitive with the best method at the other steps, indicating consistent improvements across forecasting horizons. The shaded regions further show that the observed trends are consistent across random seeds. \par
The competing methods exhibit several common performance patterns. ST-SSL and STAEformer achieve competitive results on certain datasets but show less consistent generalization across tasks. ST-SSL is primarily designed for one-step traffic forecasting, whereas STAEformer jointly learns input features and adaptive embeddings without explicitly decomposing or adaptively fusing spatial relations. In contrast, AutoTCL and InfoTS show relatively strong distributional alignment when their point-wise errors are competitive. This may be attributed to the implicit distribution alignments induced by contrastive learning and ridge regression used by these methods.
DRAN instead introduces explicit spatial-distribution alignment through SFL and $\mathcal{L}_{\rm spa}$, therefore improving distributional consistency alongside reduced prediction errors. \par
\textbf{Prediction visualization:} 
Fig. \ref{weather} shows that, while the numerical metrics are very close, DRAN produces fewer large-error regions than the strongest competing methods, indicating more spatially consistent predictions.
Additional node-level temporal prediction cases are provided in Appendix~\ref{app:temporal_vis}.\par
\textbf{Computational cost:} To evaluate the trade-off between computational efficiency and prediction accuracy, we compare inference time, memory usage, and MAE in Fig.~\ref{infer}. For AutoTCL and InfoTS, the reported inference time includes both the average ridge regression update time and the test-time forward inference. 
Although DRAN contains more parameters, it achieves the lowest MAE with competitive inference time, benefiting from the parallel computation of its temporal and spatial modules. \par
\subsection{Robustness Analysis}
To evaluate robustness under non-stationary and perturbed
conditions, we conduct stress tests on Weather and PeMS08,
which represent weather and traffic forecasting tasks, respectively.
All models are trained on the original clean training sets and
directly evaluated without retraining. We consider Gaussian sensor
noise with standard deviations $\sigma\in\{0.1,0.2,0.3\}$ in the
normalized observation space, and randomly mask
$\rho\in\{10\%,20\%,30\%\}$ of the sensors throughout the entire
lookback window. We further evaluate the models on the top 10\% most
dynamic test samples, selected according to the mean absolute
first-order difference of the forecasting targets. \par
As shown in Table~\ref{tab:robustness}, all models experience
performance degradation under increasingly severe perturbations.
Nevertheless, DRAN consistently achieves lower MAE than the
strongest competing baseline across all evaluated conditions.
Under Gaussian noise with $\sigma=0.3$, DRAN reduces MAE by
10.9\% and 9.1\% over RGSL and STAEformer on Weather and
PeMS08, respectively. Under 30\% missing sensors, the corresponding
reductions are 1.3\% and 6.8\%. These results indicate that DRAN
maintains a consistent relative advantage under severe test-time
corruptions. Detailed corruption procedures and the selection of high-dynamic
samples are provided in Appendix~\ref{app:robustness}. \par
 \subsection{Preservation of Spatial Distribution}
\begin{figure*}[htbp]
\centering
\includegraphics[width=0.72\textwidth]{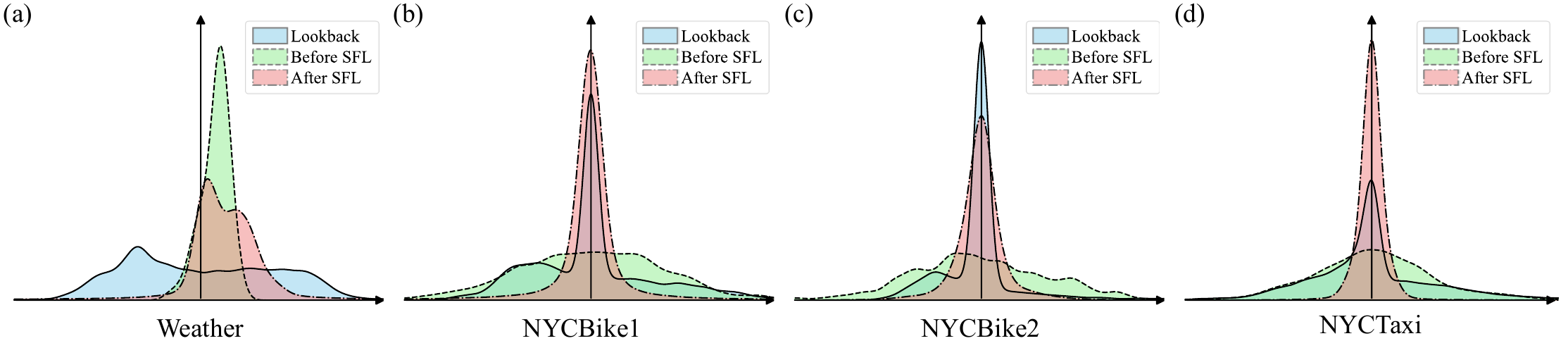}
\caption{\label{dis}Comparison of spatial distributions before and after SFL. Panels (a)–(d) correspond to the Weather, NYCBike1, NYCBike2, and NYCTaxi datasets, respectively. Each panel shows Gaussian kernel density estimates \cite{silverman2018density} for the raw inputs $\bm{X}$, the latent representations before SFL $\bm{H}_{\rm tem}$, and the representations after SFL $\bm{H}_{\rm spa}$, denoted as ``Lookback'', ``Before SFL'', and ``After SFL'', respectively.}
\end{figure*}

To evaluate the effectiveness of SFL, we aim to answer the following questions: 1) Does SFL preserve spatial distributions across various tasks? 2) Does SFL generalize across various temporal normalization strategies, performing better than feature-level normalization and other spatial normalization methods? 
To answer the first question, we analyze the spatial distributions of the raw input $\bm{X}$, the representations before SFL $\bm{H}_{\rm tem}$ and after SFL $\bm{H}_{\rm spa}$. At each time step, we construct spatial relation matrices from $\bm{X}$, $\bm{H}_{\rm tem}$, and $\bm{H}_{\rm spa}$ using pairwise $\ell_1$ distances between nodes, and evaluate the spatial discrepancies of $\bm{H}_{\rm tem}$ and $\bm{H}_{\rm spa}$ relative to $\bm{X}$ using Pairwise Distance Distortion (PDD) and 
Wasserstein Distance (${\rm WD}_{rel}$). PDD measures the distortion of node-pair spatial relations, whereas WD quantifies the distributional shift of these relations. Detailed definitions are provided in Appendix~\ref{metrics}. As shown in Table~\ref{tab:spatial_preservation}, SFL reduces PDD and WD on all datasets, demonstrating that it generally mitigates the spatial distortion introduced by temporal normalization. For qualitative analysis, we randomly sample instances from each task and estimate their spatial distributions using Gaussian kernel density estimation. As shown in Fig.~\ref{dis}, the distributions after SFL (pink) generally overlap more closely with those of the raw input (blue) than do the representations before SFL (green). Together with the quantitative results, these observations demonstrate that SFL 
better preserves the original spatial relationships and restores the spatial statistical structure distorted by temporal normalization.\par
Furthermore, to evaluate SFL’s effectiveness across different temporal normalization strategies, we replace the original normalization modules with those in DAIN, the Non-stationary Transformer, Dish-TS, and ST-Norm. SFL utilizes the mean and standard deviation of lookback windows to generate spatial factors. RevIN is included as a feature-level normalization baseline. 
For ST-Norm, we follow its original design by concatenating the temporally normalized, spatially normalized, and raw input features, and use the same downstream architecture for the variants with and without SFL.
Since the learned representations are fused with both temporally and spatially normalized components, this setup allows us to examine the effect of SFL under mixed feature compositions. Detailed implementation settings are provided in Appendix~\ref{app:norm_impl}.\par
As shown in Table \ref{dis-tab}, SFL improves the performance of various temporal normalization methods, demonstrating its generality in preserving spatial distributions. The models incorporating temporal operations and SFL outperform the model with RevIN. This finding suggests that the feature-wise normalization alone is too coarse-grained and is insufficient for distribution adaptation in spatio-temporal tasks. 
ST-Norm also benefits from SFL. However, the magnitude of improvement is smaller compared to other methods. This is likely because ST-Norm already integrates both temporally and spatially normalized features, whereas SFL mainly facilitates the recovery of temporally normalized components, providing limited additional benefit to the spatially normalized ones.
Therefore, applying SFL after temporal normalization provides an effective approach for distribution adaptation compared with instance-level, spatial-level and temporal-level normalization.\par 
\begin{figure*}[htbp]
\centering
\includegraphics[width=0.72\linewidth]{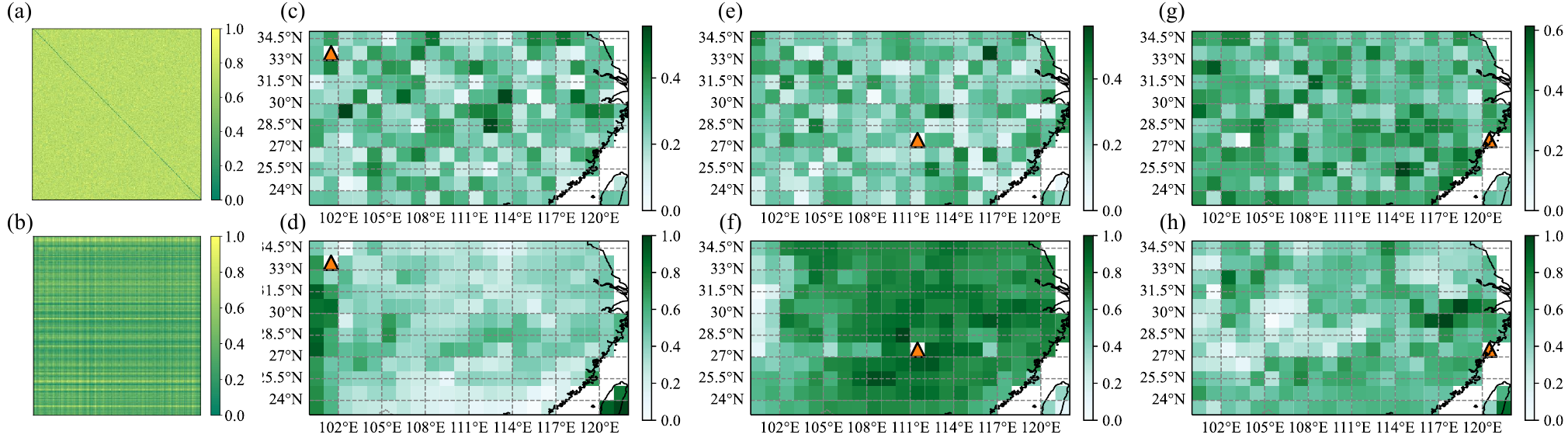}
\caption{\label{relation}Visualization of static and dynamic relations learned by DSFL on the Weather dataset. Panels (a) and (b) are the adjacency matrices of static and dynamic branches. Panels (c)–(h) show the relation strengths for three nodes located in different areas of Weather dataset. Panels (c), (e), and (g) show the static relations of the nodes, while panels (d), (f) and (h) display the dynamic relations. The orange triangles represent the selected target nodes. The darker color indicates a closer relationship between nodes. }
\end{figure*}
\subsection{Dynamic and Static Relations Learning}
In Fig. \ref{relation}, we illustrate the decomposition and fusion process of our DSFL module. Fig. \ref{relation}(a) and (b) visualize the learned static and dynamic adjacency matrices on the Weather dataset, where darker colors indicate stronger inter-node relationships. The dynamic adjacency matrix $\bm{A}_{\rm Dy}={\rm Softmax}({\bm{z}_{\rm Dy}W_{\rm Q}^{(l)}(\bm{z}_{\rm Dy}W_{\rm K}^{(l)})^{T}}/{\sqrt{C'}})$ is derived from the similarity between node time series  according to Eq. (\ref{Dy}), while the static adjacency matrix is obtained by learning the static relations $\bm{A}_{\rm St}$ according to Eq. (\ref{ast}). 
The dynamic adjacency matrix emphasizes local relations among nodes with similar signals, whereas the static matrix captures broader non-local dependencies, yielding complementary spatial representations. \par
To clarify the differences between the learned dynamic and static relations for specific nodes, we select node $i$ from various locations and visualize the relationships between the selected node and other nodes, represented as $\bm{A}_{\rm{St},\textit{i}}$ and $\bm{A}_{\rm{Dy},\textit{i}}$. As shown in Fig. \ref{relation} (c)-(h), the dynamic relations are concentrated around the target nodes, highlighting the significance of local connections. In contrast, the static relations capture interactions with distant nodes, emphasizing non-local relationships. This demonstrates that DSFL effectively learns comprehensive and complementary spatial relations.\par

\begin{figure}[htbp]
\centering
\includegraphics[width=0.75\linewidth]{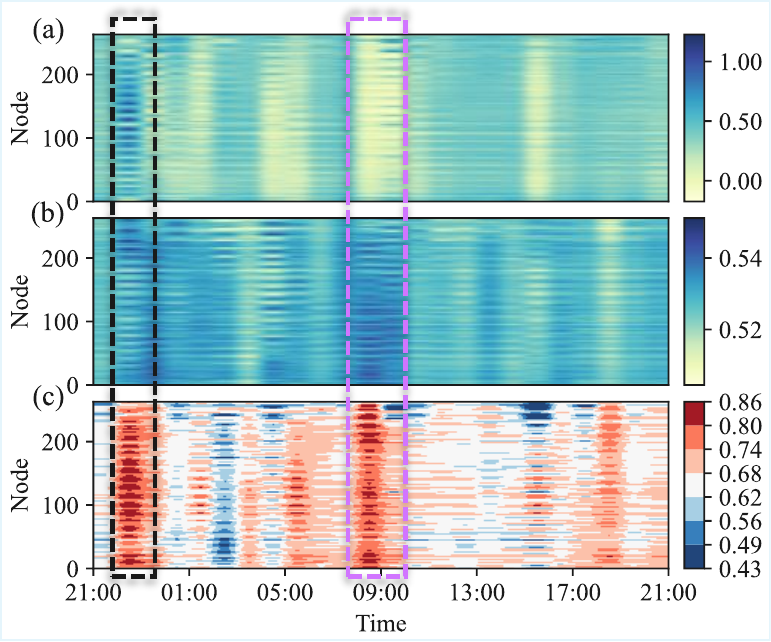}
\caption{\label{gate}A case study of the gate fusion mechanism in the DSFL module. A time window from 21:00 on 25/09/2022 to 21:00 on 26/09/2022 in the Weather dataset is selected. Panels (a) and (b) show the learned static component $\bm{H}_{\rm St}$ and dynamic component $\bm{H}_{\rm Dy}$, respectively, while (c) presents the corresponding gate signal. The highlighted regions (black and purple boxes) indicate periods of significant dynamic variation.}
\end{figure}

Furthermore, we present a case study to illustrate how the gate 
adaptively integrates static and dynamic features. We select a 
temperature window from 21:00 on 25/09/2022 to 21:00 on 26/09/2022 
from the Weather dataset. Fig.~\ref{gate} visualizes the learned static 
component $\bm{H}_{\rm St}$, dynamic component $\bm{H}_{\rm Dy}$, and 
the corresponding gate signal $z_{\rm control}$. Two prominent peaks 
in the dynamic component, highlighted by the black and purple boxes, 
coincide with large gate responses, indicating that the gate emphasizes 
periods of strong dynamic variation. These peaks occur during 
22:00$\sim$23:00 and 08:00$\sim$10:00, corresponding to rapid 
temperature decreases and increases, respectively. In contrast, the 
static component remains comparatively smooth. During 
08:00$\sim$10:00, its relatively small magnitude indicates that the 
fusion is dominated by the dynamic component. Although the static 
component also shows elevated responses during 22:00$\sim$23:00, the 
gate still assigns greater weight to the dynamic component, enabling 
adaptive fusion according to temporal variability.\par
\begin{table*}[!t]
\centering
\caption{\label{abl}The ablation results.}
\setlength{\tabcolsep}{0.5pt}
\renewcommand{\arraystretch}{0.8} 
\footnotesize
\begin{tabular}{@{}c|cc|cc|cc|cc|cc|cc@{}}
\toprule
\multirow{2}{*}{Strategies} & \multicolumn{2}{c|}{Weather}                                                                                                   & \multicolumn{2}{c|}{NYCBike1}     & \multicolumn{2}{c|}{NYCBike2}    & \multicolumn{2}{c|}{NYCTaxi}     & \multicolumn{2}{c|}{PeMS04}  & \multicolumn{2}{c}{PeMS08}                   \\  \cmidrule(lr){2-3} \cmidrule(lr){4-5}\cmidrule(lr){6-7} \cmidrule(lr){8-9} \cmidrule(lr){10-11} \cmidrule(lr){12-13}
        & MAE$\downarrow$   & WD$\downarrow$   & MAE$\downarrow$   & WD$\downarrow$ & MAE$\downarrow$   & WD$\downarrow$ & MAE$\downarrow$   & WD$\downarrow$ & MAE$\downarrow$   & WD$\downarrow$ & MAE$\downarrow$   & WD$\downarrow$ \\ \midrule
DRAN &\textbf{0.676\std{0.005}}	&\textbf{0.392\std{0.011}}	&\textbf{5.046\std{0.141}}	&\textbf{0.415\std{0.148}}	&\textbf{4.845\std{0.203}}	&\textbf{0.437\std{0.039}}	&\textbf{10.721\std{0.980}}	&\textbf{0.750\std{0.146}}	&\textbf{18.132\std{0.008}}	&\textbf{4.687\std{0.187}}	&\textbf{13.366\std{0.117}}	&\textbf{4.180\std{0.075}} \\ \midrule
w/o SFL \& $\bm{\mathcal{L}_{\rm spa}}$ &1.194\std{0.002} 	&2.000\std{0.045}	&5.502\std{0.163}	&0.720\std{0.061}	&5.266\std{0.229}	&1.495\std{0.021}	&12.426\std{1.016}	&1.520\std{0.076}	&18.642\std{0.064}	&5.712\std{0.099}	&13.995\std{0.555}	&4.822\std{0.265} \\
w/o $\bm{\mathcal{L}_{\rm spa}}$ &0.887\std{0.004}	&1.168\std{0.069}	&5.403\std{0.185}	&0.688\std{0.080}	&5.148\std{0.182}	&0.696\std{0.007}	&11.451\std{1.231}	&1.371\std{0.118}	&18.281\std{0.046}	&5.485\std{0.186}	&13.719\std{0.301}	&4.536\std{0.076}\\ \midrule
w/o DSFL	&1.193\std{0.002}	&1.542\std{0.072}	&5.529\std{0.170}	&0.730\std{0.080}	&5.536\std{0.221}	&1.577\std{0.058}	&12.276\std{1.404}	&1.612\std{0.143}	&18.695\std{0.131}	&5.586\std{0.157}	&13.663\std{0.142}	&4.613\std{0.126}\\
w/o Decomposition	&0.794\std{0.013}	&0.669\std{0.038}	&5.404\std{0.170}	&0.655\std{0.071}	&5.258\std{0.193}	&0.600\std{0.058}	&11.375\std{1.348}	&1.345\std{0.098}	&18.275\std{0.070}	&4.908\std{0.145}	&13.537\std{0.084}	&4.534\std{0.069} \\
w/o Gate &0.900\std{0.012}	&0.840\std{0.051}	&5.475\std{0.176}	&0.644\std{0.058}	&5.331\std{0.025}	&0.792\std{0.068}	&11.471\std{1.258}	&1.269\std{0.079}	&18.261\std{0.050}	&4.806\std{0.165}	&13.600\std{0.246}	&4.578\std{0.081} \\ \midrule
w/o Stochastic learner &1.009\std{0.015} &1.437\std{0.036} &5.394\std{0.125} &0.648\std{0.059} &5.211\std{0.149} &0.819\std{0.098} &11.248\std{1.164} &1.005\std{0.107} &18.319\std{0.041} &5.134\std{0.149} &13.642\std{0.461} &4.540\std{0.075}\\
\bottomrule
\end{tabular}
\end{table*}
\subsection{Ablation Studies}
To evaluate the effectiveness of each module of DRAN, we conduct an ablation study by systematically removing key components. The ablation variants are detailed as follows:
\begin{itemize}
\item \textbf{w/o SFL \& $\mathcal{L}_{\rm spa}$}: We remove the SFL module and its constraint $\mathcal{L}_{\rm spa}$, resulting in a temporal-only normalization.
\item \textbf{w/o} $\bm{\mathcal{L}_{\rm spa}}$: We remove the spatial constraint $\mathcal{L}_{\rm spa}$, eliminating the node-wise reconstruction constraint. 
\item \textbf{w/o DSFL}: We remove the DSFL module and replace it with spatial attention. 
\item \textbf{w/o Decomposition}: We remove the spatial frequency decomposition module and feed $\bm{H}_{\rm spa}$ directly into the static and dynamic branches, which only learn spatial relations without explicit frequency decomposition. 
\item \textbf{w/o Gate}: We replace the gate mechanism conditioned by static representation and input-variability statistics with a linear layer that maps concatenated features $[\bm{H}_{\rm St}, \bm{H}_{\rm Dy}]$ to the shape $\mathbb{R}^{L\times N \times C'}$.
\item \textbf{w/o Stochastic learner}: We remove the stochastic learner and use only $\bm{H}$ as the input to the decoder. 
\end{itemize}
The ablation results in Table \ref{abl} show that the model incorporating all modules achieves the best overall performance. Both SFL and DSFL contribute significantly to the final results. Removing SFL leads to performance degradation across all datasets, particularly on the WD metric, indicating that SFL plays an important role in distribution alignment and also benefits point prediction accuracy. The additional constraint $\mathcal{L}_{\rm spa}$ further facilitates spatial relation learning, thereby improving the modeling of the prediction distribution.
Furthermore, the DSFL module effectively captures both dynamic and static spatial relations. Removing the entire DSFL module results in a substantial decrease in prediction accuracy. In addition, the decomposition mechanism and the gate fusion mechanism are both critical components of DSFL. Removing either of them leads to noticeable performance degradation (DRAN vs. w/o decomposition and DRAN vs. w/o gate). The stochastic learner also contributes positively to the overall prediction performance.\par

\section{\label{sec:5}Discussion and Conclusions}
In this work, we proposed the DRAN framework to address key challenges in spatio-temporal forecasting, including distribution shifts and relational variations. DRAN integrates a Spatial Factor Learner (SFL) for distribution adaptation and a Dynamic-Static Fusion Learner (DSFL) to model both dynamic and static spatial dependencies. Experiments across weather and traffic datasets demonstrate the effectiveness and generality of both modules.\par
Despite its strong performance, DRAN shows two limitations:
\begin{itemize}
    \item \textbf{Scalability.}
    Although SFL scales linearly with the number of nodes, the dense graph convolution and spatial attention in DSFL result in an online inference complexity dominated by $O((k+2)LN^2C')$. 
    This complexity is of the same order as standard spatial self-attention and might limit scalability to very large graphs. Future work will explore sparse or low-rank spatial operators to reduce this overhead.
    \item \textbf{Adaptation to abrupt spatial changes.}
    Although DRAN remains robust to sensor noise and missing sensors, these evaluations do not explicitly cover abrupt topology changes or rare events that create previously unseen spatial relations. Because the static branch is learned from training data and the dynamic branch relies on a finite lookback window, DSFL may adapt slowly in such cases. Future work will investigate event-aware online updates of spatial relations and fusion weights.
\end{itemize}

\bibliographystyle{IEEEtran}
\bibliography{IEEEabrv,main}

@STRING{IEEE_J_ITS        = "{IEEE} Trans. Intell. Transp. Syst."}

@STRING{IEEE_J_CASI_RP    = "{IEEE} Trans. Circuits Syst. {I}"}

@STRING{IEEE_J_NNLS       = "{IEEE} Trans. Neural Netw. Learn. Syst."}

@STRING{IEEE_J_CYB        = "{IEEE} Trans. Cybern."}

@STRING{CHAOS = "Chaos"}

@STRING{ERA = "Electron. Res. Arch."}

@STRING{PHYSREP = "Phys. Rep."}

@STRING{AAAI = "Proc. AAAI Conf. Artif. Intell."}

@STRING{NEURIPS = "Proc. Adv. Neural Inf. Process. Syst."}

@article{FWBNet,
title = {Frequency-wavelet adaptive basis network for long-term time series forecasting},
journal = {Eng. Appl. Artif. Intell.},
volume = {161},
pages = {112161},
year = {2025},
issn = {0952--1976},
doi = {10.1016/j.engappai.2025.112161},
author = {Qiang Lai and Yang You},
}

@inproceedings{decomp_freq,
  author={Ma, Jingtian and Wang, Jingyuan and U, Leong Hou},
  title={Hierarchical Frequency-Decomposition Graph Neural Networks
         for Road Network Representation Learning},
  booktitle={Proc. AAAI Conf. Artif. Intell.},
  year={2026},
  volume={40},
  number={18},
  pages={15510--15518},
  doi={10.1609/aaai.v40i18.38579}
}

@article{MixGT,
title = {Unveiling node relationships for traffic forecasting: A self-supervised approach with MixGT},
journal = {Inf. Fusion},
volume = {120},
pages = {103070},
year = {2025},
issn = {1566--2535},
doi = {10.1016/j.inffus.2025.103070},
author = {Qiang Lai and Peng Chen},
}

@inproceedings{decomp_direct,
author = {Wu, Zonghan and Pan, Shirui and Long, Guodong and Jiang, Jing and Chang, Xiaojun and Zhang, Chengqi},
title = {Connecting the Dots: Multivariate Time Series Forecasting with Graph Neural Networks},
year = {2020},
doi = {10.1145/3394486.3403118},
booktitle = {Proc. 26th ACM SIGKDD Int. Conf. Knowl. Discov. Data Min.},
pages = {753--763},
numpages = {11},
}

@article{LEISN,
title = {LEISN: A long explicit–implicit spatio-temporal network for traffic flow forecasting},
journal = {Expert Syst. Appl.},
volume = {245},
pages = {123139},
year = {2024},
issn = {0957--4174},
doi = {10.1016/j.eswa.2024.123139},
author = {Qiang Lai and Peng Chen},
}

@inproceedings{decomp_spec,
author = {Geisler, Simon and Kosmala, Arthur and Herbst, Daniel and G\"{u}nnemann, Stephan},
title = {Spatio-spectral graph neural networks},
year = {2024},
isbn = {9798331314385},
booktitle = NEURIPS,
articleno = {1554},
numpages = {59},
}

@article{nt2019revisiting,
  title={Revisiting graph neural networks: All we have is low-pass filters},
  author={{NT}, Hoang and Maehara, Takanori},
  journal={arXiv preprint arXiv:1905.09550},
  year={2019}
}

@book{silverman2018density,
  title={Density estimation for statistics and data analysis},
  author={Silverman, Bernard W},
  year={2018},
  publisher={Routledge},
address   = {Boca Raton, FL, USA},
}

@article{samsgl,
    author = {Zou, Xiaobei and Xiong, Luolin and Tang, Yang and Kurths, Jürgen},
    title = {{SAMSGL}: Series-aligned multi-scale graph learning for spatiotemporal forecasting},
    journal = CHAOS ,
    volume = {34},
    number = {6},
    pages = {063140},
    year = {2024},
    month = jun,
    issn = {1054--1500},
    doi = {10.1063/5.0211403},
}

@ARTICLE{10491369,
  author={Perrusquía, Adolfo and Guo, Weisi},
  journal=IEEE_J_CYB, 
  title={Reservoir Computing for Drone Trajectory Intent Prediction: A Physics Informed Approach}, 
  year={2024},
  volume={54},
  number={9},
  pages={4939--4948},
}

@ARTICLE{9457154,
  author={Liu, Yiqun and Zhang, Junping and Chen, Lei and Chu, Hai and Wang, James Z. and Ma, Leiming},
  journal=IEEE_J_CYB, 
  title={{SSAS}: Spatiotemporal Scale Adaptive Selection for Improving Bias Correction on Precipitation}, 
  year={2022},
  volume={52},
  number={11},
  pages={12175--12188},
}

@ARTICLE{9983531,
  author={Pu, Bin and Liu, Jiansong and Kang, Yan and Chen, Jianguo and Yu, Philip S.},
  journal=IEEE_J_CYB, 
  title={{MVSTT}: A Multiview Spatial-Temporal Transformer Network for Traffic-Flow Forecasting}, 
  year={2024},
  volume={54},
  number={3},
  pages={1582--1595},
  doi={10.1109/TCYB.2022.3223918}}

@inproceedings{
lee2024testam,
title={{TESTAM}: A Time-Enhanced Spatio-Temporal Attention Model with Mixture of Experts},
author={Hyunwook Lee and Sungahn Ko},
booktitle={Proc. Int. Conf. Learn. Representations},
year={2024},
}

@inproceedings{stnorm,
author = {Deng, Jinliang and Chen, Xiusi and Jiang, Renhe and Song, Xuan and Tsang, Ivor W.},
title = {{ST}-{N}orm: Spatial and Temporal Normalization for Multi-variate Time Series Forecasting},
year = {2021},
isbn = {9781450383325},
doi = {10.1145/3447548.3467330},
booktitle = {Proc. 27th ACM SIGKDD Conf. Knowl. Discov. Data Mining},
pages = {269--278},
numpages = {10},
}

@inproceedings{memda,
author = {Cai, Zekun and Jiang, Renhe and Yang, Xinyu and Wang, Zhaonan and Guo, Diansheng and Kobayashi, Hill Hiroki and Song, Xuan and Shibasaki, Ryosuke},
title = {MemDA: Forecasting Urban Time Series with Memory-based Drift Adaptation},
year = {2023},
doi = {10.1145/3583780.3614962},
pages = {193--202},
numpages = {10},
booktitle= {Proc. ACM Int. Conf. Inf. Knowl. Manage.}
}

@inproceedings{
parats,
title={Parametric Augmentation for Time Series Contrastive Learning},
author={Xu Zheng and Tianchun Wang and Wei Cheng and Aitian Ma and Haifeng Chen and Mo Sha and Dongsheng Luo},
booktitle={Proc. Int. Conf. Learn. Representations},
year={2024},
}

@inproceedings{infots, title={Time Series Contrastive Learning with Information-Aware Augmentations}, volume={37}, 
DOI={10.1609/aaai.v37i4.25575},  number={4}, booktitle={Proc. AAAI Conf. Artif. Intell.}, author={Luo, Dongsheng and Cheng, Wei and Wang, Yingheng and Xu, Dongkuan and Ni, Jingchao and Yu, Wenchao and Zhang, Xuchao and Liu, Yanchi and Chen, Yuncong and Chen, Haifeng and Zhang, Xiang}, year={2023}, month=jun, pages={4534--4542} }

@inproceedings{dualts,
author = {Qin, Yao and Song, Dongjin and Cheng, Haifeng and Cheng, Wei and Jiang, Guofei and Cottrell, Garrison W.},
title = {A dual-stage attention-based recurrent neural network for time series prediction},
year = {2017},
isbn = {9780999241103},
booktitle = {Proc. 26th Int. Joint Conf. Artif. Intell.},
pages = {2627--2633},
numpages = {7},
}

@article{ga2022,
  title={Deep learning for time series forecasting: The electric load case},
  author={Gasparin, Alberto and Lukovic, Slobodan and Alippi, Cesare},
  journal={CAAI Trans. Intell. Technol.},
  volume={7},
  number={1},
  pages={1--25},
  year={2022},
  publisher={Wiley Online Library}
}

@article{xiong2022two,
  author={Xiong, Luolin and Tang, Yang and Mao, Shuai and Liu, Hangyue and Meng, Ke and Dong, Zhaoyang and Qian, Feng},
  journal=IEEE_J_CASI_RP, 
  title={A Two-Level Energy Management Strategy for Multi-Microgrid Systems With Interval Prediction and Reinforcement Learning}, 
  year={2022},
  volume={69},
  number={4},
  pages={1788--1799},
  doi={10.1109/TCSI.2022.3141229}}

@ARTICLE{9764831,
  author={Tang, Yang and Zhao, Chaoqiang and Wang, Jianrui and Zhang, Chongzhen and Sun, Qiyu and Zheng, Wei Xing and Du, Wenli and Qian, Feng and Kurths, Jürgen},
  journal=IEEE_J_NNLS, 
  title={Perception and Navigation in Autonomous Systems in the Era of Learning: A Survey}, 
  year={2023},
  volume={34},
  number={12},
  pages={9604--9624},
  doi={10.1109/TNNLS.2022.3167688}}

@inproceedings{Cini_Alippi_2023, 
title={Scalable Spatiotemporal Graph Neural Networks}, 
volume={37}, 
DOI={10.1609/aaai.v37i6.25880}, 
booktitle=AAAI,
author={Cini, Andrea and Marisca, Ivan and Bianchi, Filippo Maria and Alippi, Cesare}, 
year={2023}, 
month=jun, 
pages={7218--7226} }

@inproceedings{ji2023spatio, 
title={Spatio-Temporal Self-Supervised Learning for Traffic Flow Prediction}, volume={37}, 
DOI={10.1609/aaai.v37i4.25555},  
number={4}, 
booktitle=AAAI, 
author={Ji, Jiahao and Wang, Jingyuan and Huang, Chao and Wu, Junjie and Xu, Boren and Wu, Zhenhe and Zhang, Junbo and Zheng, Yu}, 
year={2023}, 
month=jun, 
pages={4356--4364} }

@article{era5,
author = {Hersbach, Hans and Bell, Bill and Berrisford, Paul and Hirahara, Shoji and Horányi, András and Muñoz-Sabater, Joaquín and Nicolas, Julien and Peubey, Carole and Radu, Raluca and Schepers, Dinand and Simmons, Adrian and Soci, Cornel and Abdalla, Saleh and Abellan, Xavier and Balsamo, Gianpaolo and Bechtold, Peter and Biavati, Gionata and Bidlot, Jean and Bonavita, Massimo and De Chiara, Giovanna and Dahlgren, Per and Dee, Dick and Diamantakis, Michail and Dragani, Rossana and Flemming, Johannes and Forbes, Richard and Fuentes, Manuel and Geer, Alan and Haimberger, Leo and Healy, Sean and Hogan, Robin J. and Hólm, Elías and Janisková, Marta and Keeley, Sarah and Laloyaux, Patrick and Lopez, Philippe and Lupu, Cristina and Radnoti, Gabor and de Rosnay, Patricia and Rozum, Iryna and Vamborg, Freja and Villaume, Sebastien and Thépaut, Jean-Noël},
title = {The ERA5 global reanalysis},
journal = {Quart. J. Roy. Meteorol. Soc.},
volume = {146},
number = {730},
pages = {1999--2049},
doi = {10.1002/qj.3803},
year = {2020},
month     = may,
}

@inproceedings{liu2023spatio,
    author = {Liu, Hangchen and Dong, Zheng and Jiang, Renhe and Deng, Jiewen and Deng, Jinliang and Chen, Quanjun and Song, Xuan},
  title = {Spatio-Temporal Adaptive Embedding Makes Vanilla Transformer SOTA for Traffic Forecasting},
year = {2023},
doi = {10.1145/3583780.3615160},
booktitle = {Proc. ACM Int. Conf. Inf. Knowl. Manage.},
pages = {4125--4129},
numpages = {5},
}

@inproceedings{seo2018structured,
  author="Seo, Youngjoo
and Defferrard, Micha{\"e}l
and Vandergheynst, Pierre
and Bresson, Xavier",
editor="Cheng, Long
and Leung, Andrew Chi Sing
and Ozawa, Seiichi",
title="Structured Sequence Modeling with Graph Convolutional Recurrent Networks",
booktitle="Int. Conf. Neural Inf. Process.",
year="2018",
publisher="Springer International Publishing",
address="Cham",
pages="362--373",

}

@article{wu2022survey,
  title = {A survey on temporal network dynamics with incomplete data},
journal = {Electron. Res. Arch.},
volume = {30},
number = {10},
pages = {3786--3810},
year = {2022},
issn = {2688-1594},
doi = {10.3934/era.2022193},
author = {Xing Wu and Shuai Mao and Luolin Xiong and Yang Tang},
}

@inproceedings{ijcai2022p328,
  title     = {Regularized Graph Structure Learning with Semantic Knowledge for Multi-variates Time-Series Forecasting},
  author    = {Yu, Hongyuan and Li, Ting and Yu, Weichen and Li, Jianguo and Huang, Yan and Wang, Liang and Liu, Alex},
  booktitle = {Proc. 31st Int. Joint Conf. Artif. Intell.},
  pages     = {2362--2368},
  year      = {2022},
  month     = jul,
  doi       = {10.24963/ijcai.2022/328},
}

@inproceedings{bai2020adaptive,
 author = {Bai, Lei and Yao, Lina and Li, Can and Wang, Xianzhi and Wang, Can},
 booktitle = NEURIPS ,
 pages = {17804--17815},
 title = {Adaptive Graph Convolutional Recurrent Network for Traffic Forecasting},
 volume = {33},
 year = {2020}
}

@inproceedings{Meta_Graph, title={Spatio-Temporal Meta-Graph Learning for Traffic Forecasting}, volume={37}, 
DOI={10.1609/aaai.v37i7.25976}, number={7}, booktitle=AAAI, author={Jiang, Renhe and Wang, Zhaonan and Yong, Jiawei and Jeph, Puneet and Chen, Quanjun and Kobayashi, Yasumasa and Song, Xuan and Fukushima, Shintaro and Suzumura, Toyotaro}, year={2023}, month=jun, pages={8078--8086} }

@article{li2023dynamic,
author = {Li, Fuxian and Feng, Jie and Yan, Huan and Jin, Guangyin and Yang, Fan and Sun, Funing and Jin, Depeng and Li, Yong},
title = {Dynamic Graph Convolutional Recurrent Network for Traffic Prediction: Benchmark and Solution},
year = {2023},
issue_date = {January 2023},
publisher = {Association for Computing Machinery},
address = {New York, NY, USA},
volume = {17},
number = {1},
issn = {1556--4681},
doi = {10.1145/3532611},
journal = {ACM Trans. Knowl. Discov. Data},
month = feb,
articleno = {9},
numpages = {21},
}

@article{ji2023signal,
  title={Signal propagation in complex networks},
  author={Ji, Peng and Ye, Jiachen and Mu, Yu and Lin, Wei and Tian, Yang and Hens, Chittaranjan and Perc, Matja{\v{z}} and Tang, Yang and Sun, Jie and Kurths, J{\"u}rgen},
  journal=PHYSREP,
  volume={1017},
  pages={1--96},
  year={2023},
  publisher={Elsevier}
}

@inproceedings{fang2023learning,
title={Learning Decomposed Spatial Relations for Multi-Variate Time-Series Modeling}, 
volume={37}, 
number={6}, 
booktitle=AAAI, 
author={Fang, Yuchen and Ren, Kan and Shan, Caihua and Shen, Yifei and Li, You and Zhang, Weinan and Yu, Yong and Li, Dongsheng}, year={2023}, month=jun, 
pages={7530--7538}
}

@inproceedings{liu2022non,
 author = {Liu, Yong and Wu, Haixu and Wang, Jianmin and Long, Mingsheng},
 booktitle = NEURIPS ,
 pages = {9881--9893},
 title = {Non-stationary Transformers: Exploring the Stationarity in Time Series Forecasting},
 volume = {35},
 year = {2022}
}

@inproceedings{fan2023dish,
    title={Dish-TS: A General Paradigm for Alleviating Distribution Shift in Time Series Forecasting}, volume={37}, 
    DOI={10.1609/aaai.v37i6.25914}, number={6}, booktitle=AAAI, 
    author={Fan, Wei and Wang, Pengyang and Wang, Dongkun and Wang, Dongjie and Zhou, Yuanchun and Fu, Yanjie}, year={2023}, month=jun, pages={7522--7529}
}

@inproceedings{
kim2021reversible,
title={Reversible Instance Normalization for Accurate Time-Series Forecasting against Distribution Shift},
author={Taesung Kim and Jinhee Kim and Yunwon Tae and Cheonbok Park and Jang-Ho Choi and Jaegul Choo},
booktitle={Proc. Int. Conf. Learn. Representations},
year={2022},
}

@inproceedings{yu2018spatio,
author = {Yu, Bing and Yin, Haoteng and Zhu, Zhanxing},
title = {Spatio-temporal graph convolutional networks: a deep learning framework for traffic forecasting},
year = {2018},
booktitle = {Proc. 27th Int. Joint Conf. Artif. Intell.},
pages = {3634--3640},
numpages = {7},
location = {Stockholm, Sweden},
}

@inproceedings{stdn, 
title={Spatiotemporal-aware Trend-Seasonality Decomposition Network for Traffic Flow Forecasting}, volume={39}, 
number={11}, booktitle=AAAI, 
author={Cao, Lingxiao and Wang, Bin and Jiang, Guiyuan and Yu, Yanwei and Dong, Junyu}, year={2025}, month=apr, pages={11463--11471} }

@inproceedings{dstmamba, title={Decomposed Spatio-Temporal Mamba for Long-Term Traffic Prediction}, volume={39}, 
DOI={10.1609/aaai.v39i11.33281}, number={11}, booktitle=AAAI, author={He, Sicheng and Ji, Junzhong and Lei, Minglong}, year={2025}, month=apr, pages={11772--11780} }

@inproceedings{guo2019attention,
  title={Attention Based Spatial-Temporal Graph Convolutional Networks for Traffic Flow Forecasting}, volume={33}, 
  DOI={10.1609/aaai.v33i01.3301922}, number={01}, booktitle=AAAI, author={Guo, Shengnan and Lin, Youfang and Feng, Ning and Song, Chao and Wan, Huaiyu}, year={2019}, month=jul, pages={922--929} 
}

@inproceedings{
li2018diffusion,
title={Diffusion Convolutional Recurrent Neural Network: Data-Driven Traffic Forecasting},
author={Yaguang Li and Rose Yu and Cyrus Shahabi and Yan Liu},
booktitle={Proc. Int. Conf. Learn. Representations},
year={2018},
}

@article{zhao2019t,
author={Zhao, Ling and Song, Yujiao and Zhang, Chao and Liu, Yu and Wang, Pu and Lin, Tao and Deng, Min and Li, Haifeng},
  journal=IEEE_J_ITS, 
  title={T-{GCN}: A Temporal Graph Convolutional Network for Traffic Prediction}, 
  year={2020},
  volume={21},
  number={9},
  pages={3848--3858},
  doi={10.1109/TITS.2019.2935152}
}

@article{zhang2024caformer,
  author={Zhang, Kexuan and Zou, Xiaobei and Yen, Gary G. and Tang, Yang and Kurths, Jürgen},
  journal=IEEE_J_CYB, 
  title={Caformer: Rethinking Time-Series Forecasting From Causal Perspective}, 
  year={2026},
  volume={56},
  number={3},
  pages={1611--1624},
  doi={10.1109/TCYB.2025.3631588}}

@article{tang2020introduction,
    author = {Tang, Yang and Kurths, Jürgen and Lin, Wei and Ott, Edward and Kocarev, Ljupco},
    title = "{Introduction to Focus Issue: When machine learning meets complex systems: Networks, chaos, and nonlinear dynamics}",
    journal = CHAOS ,
    volume = {30},
    number = {6},
    pages = {063151},
    year = {2020},
    month = jun,
    issn = {1054-1500},
    doi = {10.1063/5.0016505},
    eprint = {https://pubs.aip.org/aip/cha/article-pdf/doi/10.1063/5.0016505/14630812/063151\_1\_online.pdf},
}

@article{passalis2019deep,
  title={Deep adaptive input normalization for time series forecasting},
  author={Passalis, Nikolaos and Tefas, Anastasios and Kanniainen, Juho and Gabbouj, Moncef and Iosifidis, Alexandros},
  journal=IEEE_J_NNLS,
  volume={31},
  number={9},
  pages={3760--3765},
  year={2019},
  publisher={IEEE}
}

@article{chen2001freeway,
  title={Freeway performance measurement system: mining loop detector data},
  author={Chen, Chao and Petty, Karl and Skabardonis, Alexander and Varaiya, Pravin and Jia, Zhanfeng},
  journal={Transp. Res. Rec.},
  volume={1748},
  number={1},
  pages={96--102},
  year={2001},
  publisher={SAGE Publications Sage CA: Los Angeles, CA}
}

@inproceedings{HA,
  title={Incorporating current information into historical-average-based forecasts to improve crop price basis forecasts},
    booktitle={Proc. NCR-134 Conf. Appl. Commodity Price Anal., Forecasting, Market Risk Manage.},
  author={Taylor, Mykel R and Dhuyvetter, Kevin C and Kastens, Terry L},
  year={2004},
  volume = {29},
    month = apr
}

@article{ARIMA2,
  title={Time series forecasting using a hybrid ARIMA and neural network model},
  author={Zhang, G Peter},
  journal={Neurocomputing},
  volume={50},
  pages={159--175},
  year={2003},
  publisher={Elsevier}
}

@inproceedings{rnn1,
  title={Application of a dynamic recurrent neural network in spatio-temporal forecasting},
  author={Cheng, Tao and Wang, Jiaqiu},
  booktitle={Proc. Int. Workshop Inf. Fusion Geogr. Inf. Syst.},
  pages={173--186},
  year={2007},
}

@inproceedings{
PatchTST,
title={A time series is worth 64 words:  Long-term forecasting with transformers},
author={Yuqi Nie and Nam H. Nguyen and Phanwadee Sinthong and Jayant Kalagnanam},
booktitle={Proc. Int. Conf. Learn. Representations },
year={2023},
}

@inproceedings{autoformer,
 author = {Wu, Haixu and Xu, Jiehui and Wang, Jianmin and Long, Mingsheng},
 booktitle = NEURIPS ,
 pages = {22419--22430},
 title = {Autoformer: Decomposition Transformers with Auto-Correlation for Long-Term Series Forecasting},
 volume = {34},
 year = {2021}
}

@ARTICLE{causalllm,
  author={Li, Shijie and Li, He and Li, Xiaojing and Xu, Yong and Lin, Zhenhong and Jiang, Huaiguang},
  journal=IEEE_J_CYB, 
  title={Causal Intervention Is What Large Language Models Need for Spatio-Temporal Forecasting}, 
  year={2025},
  volume={55},
  number={8},
  pages={3825--3837},
  }

@ARTICLE{dygtcyb,
  author={Liu, Rui and Hu, Yao and Wu, Jibin and Wong, Ka-Chun and Huang, Zhi-An and Huang, Yu-An and Chen Tan, Kay},
  journal=IEEE_J_CYB, 
  title={Dynamic Graph Representation Learning for Spatio-Temporal Neuroimaging Analysis}, 
  year={2025},
  volume={55},
  number={3},
  pages={1121-1134},
  doi={10.1109/TCYB.2025.3531657}}

@ARTICLE{conti_tcyb,
  author={Zhou, Ming and Lu, Jie},
  journal=IEEE_J_CYB,
  title={Continuous Graph Learning-Based Self-Adaptation for Multi-Stream Concept Drift}, 
  year={2025},
  volume={55},
  number={8},
  pages={3760--3773},
  doi={10.1109/TCYB.2025.3569816}}

@ARTICLE{multistream_tcyb,
  author={Zhang, Bin and Lu, Jie and Song, Yiliao and Zhang, Guangquan},
  journal=IEEE_J_CYB, 
  title={A Multistream Concept Drift Handling Framework via Data Sharing}, 
  year={2025},
  volume={55},
  number={12},
  pages={5989--6000},
  doi={10.1109/TCYB.2025.3605663}}

@ARTICLE{pidada,
  author={Liu, Hongwei and Liu, Ning and Yu, Wen and Li, Xiaoou
          and Li, Yousheng and Jia, Yao and Chai, Tianyou},
  journal=IEEE_J_CYB,
  title={{PID}-Optimized Deep Learning for Adaptive Time--Frequency
         Forecasting in Dynamic Systems: Coal Calorific Value Prediction},
  year={2026},
  pages={1--14},
  doi={10.1109/TCYB.2026.3667581},
  note={early access}
}

@inproceedings{apt,
author = {Li, Yujie and Shao, Zezhi and Yu, Chengqing and Fu, Yisong and Sun, Tao and Xu, Yongjun and Wang, Fei},
title = {APT: affine prototype-timestamp for time series forecasting under distribution shift},
  booktitle = {Proc. AAAI Conf. Artif. Intell.},
  volume    = {40},
  number    = {18},
  pages     = {15180--15188},
  year      = {2026},
  doi       = {10.1609/aaai.v40i18.38542}
}

@inproceedings{tafas,
author = {Kim, HyunGi and Kim, Siwon and Mok, Jisoo and Yoon, Sungroh},
title = {Battling the non-stationarity in time series forecasting via test-time adaptation},
  booktitle = {Proc. AAAI Conf. Artif. Intell.},
  volume    = {39},
  number    = {17},
  pages     = {17868--17876},
  year      = {2025},
  doi       = {10.1609/aaai.v39i17.33965}
}

@inproceedings{zou2026metadg,
  title={Meta Dynamic Graph for Traffic Flow Prediction},
  author={Zou, Yiqing and Yuan, Hanning and Yang, Qianyu and Yuan, Ziqiang and Wang, Shuliang and Ruan, Sijie},
  booktitle=AAAI,
  volume={40},
  number={19},
  pages={16584--16592},
  year={2026},
  doi={10.1609/aaai.v40i19.38699}
}

@inproceedings{qin2026adafre,
  title={Adaptive Frequency Pathways for Spatiotemporal Forecasting},
  author={Qin, Yanjun and Fang, Yuchen and Jiang, Xinke and Miao, Hao and Tao, Xiaoming},
  booktitle=AAAI,
  volume={40},
  number={18},
  pages={15653--15661},
  year={2026},
  doi={10.1609/aaai.v40i18.38595}
}

@inproceedings{redynet,
author = {Gao, Qiang and Wang, Zizheng and Huang, Li and Trajcevski, Goce and Liu, Guisong and Chen, Xueqin},
title = {Responsive dynamic graph disentanglement for metro flow forecasting},
year = {2025},
doi = {10.1609/aaai.v39i11.33272},
booktitle = AAAI,
number={11},
pages={11690--11698},
volume={39},
}

@inproceedings{HMformer,
author = {Huang, Renjun and Xiao, Han and Li, Bingqing and Zhang, Baili and Lyu, Jianhua},
title = {HMformer: unleashing transformer's potential for time series forecasting via hierarchical multi-scale modeling},
year = {2026},
url = {https://doi.org/10.1609/aaai.v40i26.39355},
doi = {10.1609/aaai.v40i26.39355},
booktitle = AAAI,
pages={22012--22020},
numpages = {9},
}

\begin{IEEEbiography}
[{\includegraphics[width=1.1in,height=1.40in,clip,keepaspectratio]{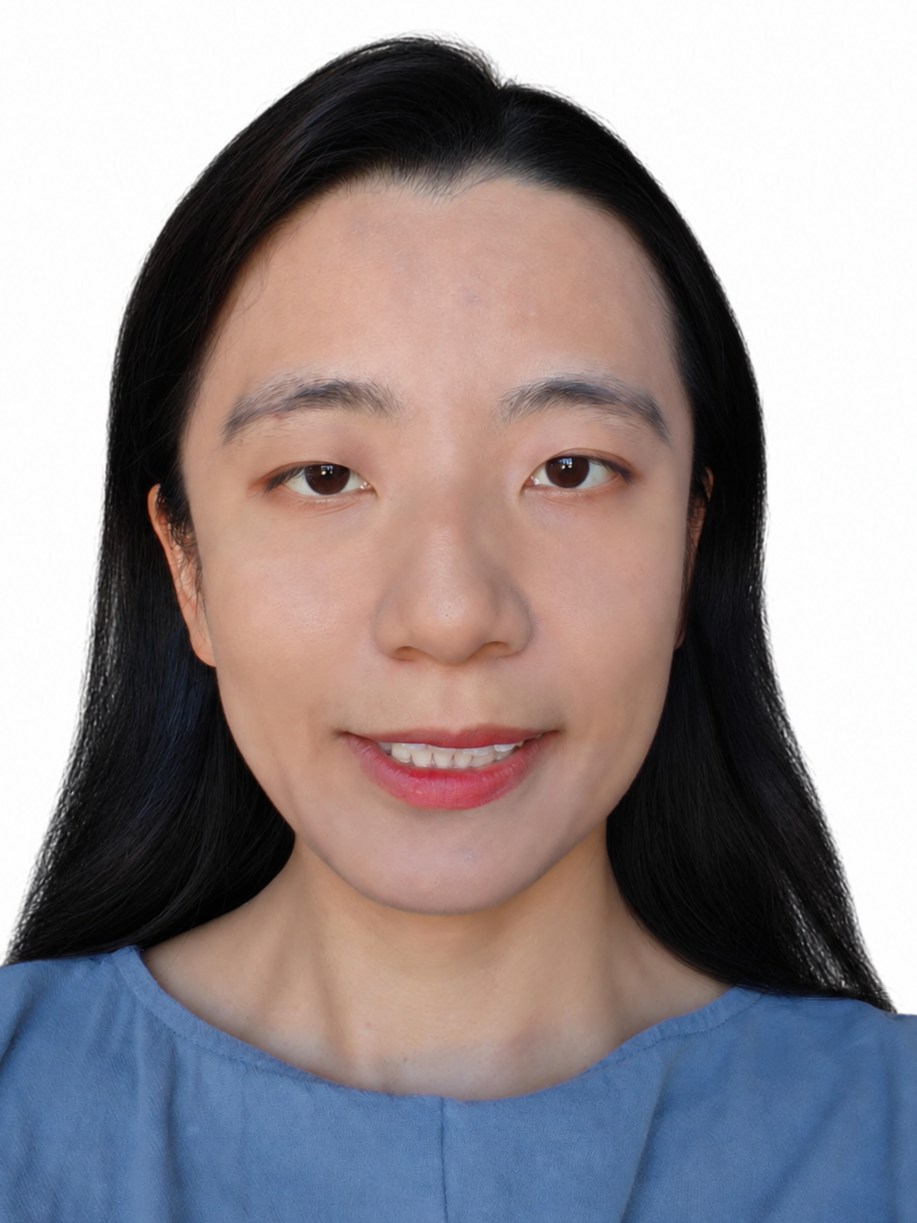}}]{Xiaobei Zou} received the B.S. degree in electrical engineering and automation from Soochow University, Suzhou, China, in 2022. She is currently pursuing the Ph.D. degree in control science and engineering at the East China University of Science and Technology, Shanghai, China. 

Her research interests include spatio-temporal forecasting, time series analysis, and graph neural networks. 
\end{IEEEbiography}

\begin{IEEEbiography}[{\includegraphics[width=1.1in,height=1.4in,clip,keepaspectratio]{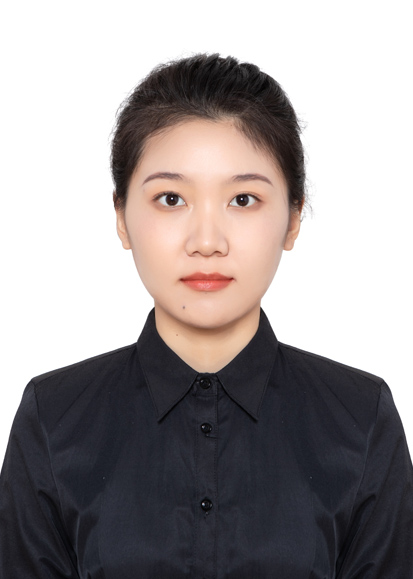}}]{Luolin Xiong}
received the B.S. and Ph.D. degree in School of Control Science and Engineering from East China University of Science and Technology, Shanghai, China, in 2020 and 2025, respectively. She is currently a Postdoctoral Researcher at East China University of Science and Technology. Her research interests include smart grid, microgrids, reinforcement learning and their applications.
\end{IEEEbiography}

\begin{IEEEbiography}
[{\includegraphics[width=1.1in,height=1.40in,clip,keepaspectratio]{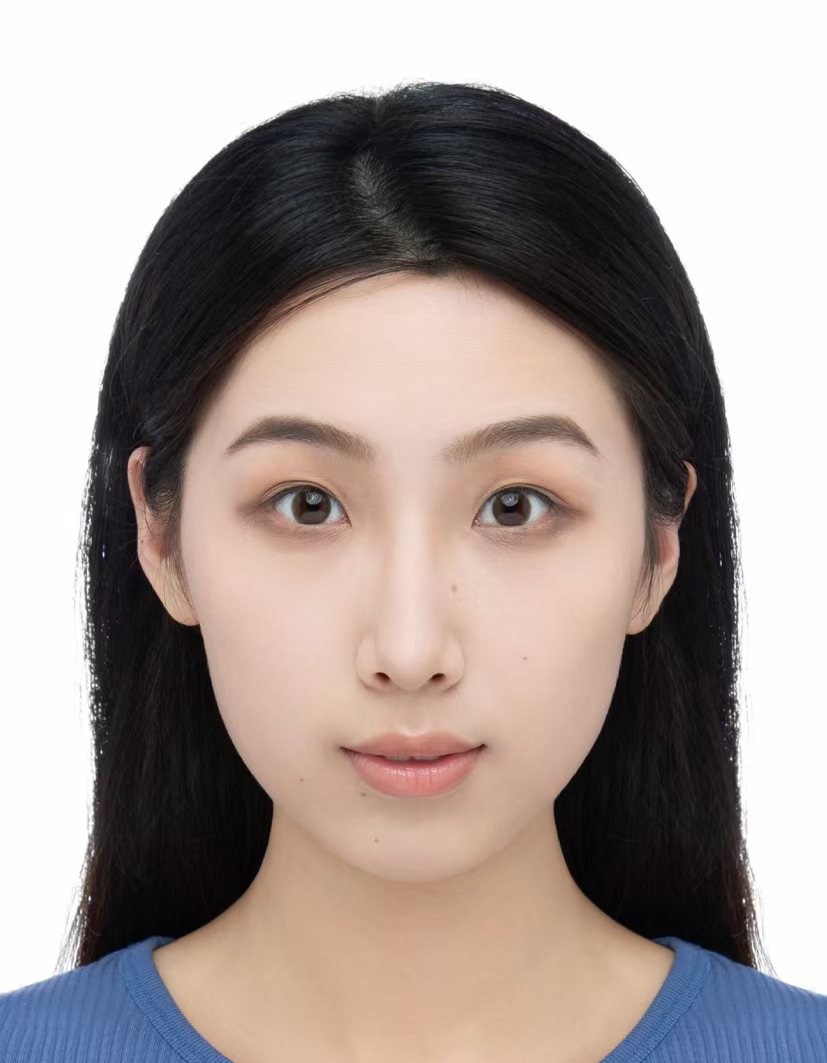}}]{Kexuan Zhang} received the B.S. degree in automation from the East China University of Science and Technology, Shanghai, China, in 2022, where she is currently pursuing the Ph.D. degree in control science and engineering. 

Her research interests include time series analysis, deep learning, and causal inference.
\end{IEEEbiography}

\begin{IEEEbiography}
[{\includegraphics[width=1.1in,height=1.40in,clip,keepaspectratio]{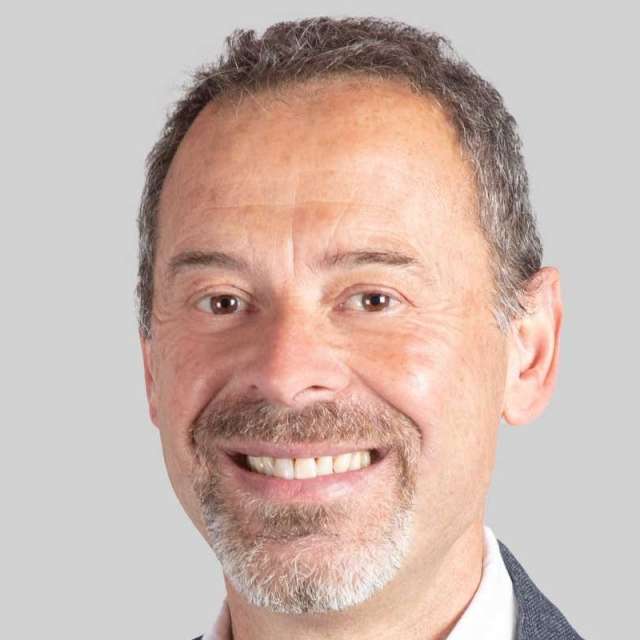}}]{Cesare Alippi}(Fellow, IEEE) is a Professor within the Faculty of Informatics at Universit`a della Svizzera italiana, professor at Politecnico di Milano, Italy, visiting professor at the Guangdong University of Technology, China and Advisory Professor at the Northwestern Polytechnic in Xi’An, China. He was a visiting researcher/professor at UCL (UK), MIT (USA), ESPCI (F), CASIA (RC), A*STAR (SIN), UKobe (JP). Alippi is an IEEE Fellow, ELLIS Fellow, INNS Fellow, AAIA Fellow, past Board of Governors
member of the International Neural Network Society; he served the IEEE CIS in several capacities, also covering the roles of Vice-President for education and chairing the Awards committee. He received several awards including the International Neural Networks Society Gabor award and the IEEE Computational Intelligence Society Enrique Ruspini meritorious
Award. 

Current research activity addresses adaptation and learning in non-stationary environments, graph learning and Intelligence for embedded, IoT and cyberphysical systems. He holds 8 patents, has published one monograph book, 7 edited books and more than 250 papers in international journals and conference proceedings.
\end{IEEEbiography}

\begin{IEEEbiography}
[{\includegraphics[width=1.1in,height=1.40in,clip,keepaspectratio]{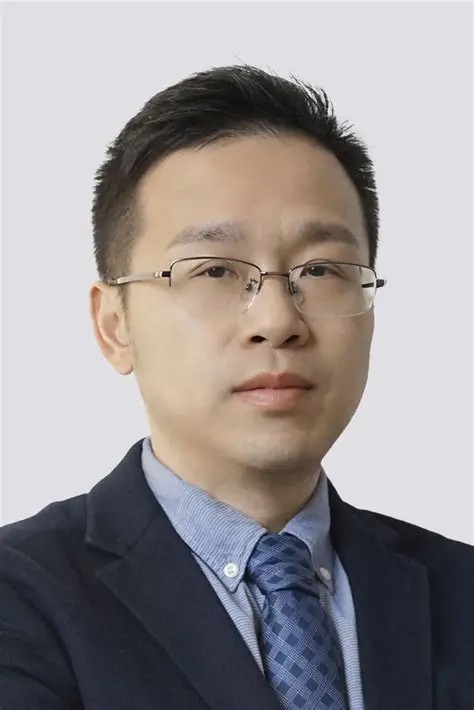}}]{Yang Tang}(Fellow, IEEE) received the B.S. and Ph.D. degrees in electrical engineering from Donghua University, Shanghai, China, in 2006 and 2010, respectively. From 2008 to 2010, he was a Research Associate with The Hong Kong Polytechnic University, Hong Kong. From 2011 to 2015, he was a Post-Doctoral Researcher with the Humboldt University of Berlin, Berlin, Germany, and with the Potsdam Institute for Climate Impact Research, Potsdam, Germany. He is currently a Professor with East China University of Science and Technology,
Shanghai. He has published more than 200 papers in international journals and conferences, including more than 160 papers in IEEE TRANSACTIONS and 20 articles in IFAC journals. His current research interests include distributed estimation /control /optimization, computer vision, reinforcement learning, cyberphysical systems, hybrid dynamical systems, and their applications. He was a recipient of the Alexander von Humboldt Fellowship. He is a Co-Editor-in-Chief of IEEE TRANSACTIONS ON INDUSTRIAL INFORMATICS, an Senior Area Editor of IEEE TRANSACTIONS ON CIRCUITS AND SYSTEMSI: REGULAR PAPERS, and an Associate Editor of IEEE TRANSACTIONS ON NEURAL NETWORKS AND LEARNING SYSTEMS, IEEE TRANSACTIONS ON CYBERNETICS, IEEE TRANSACTIONS ON INDUSTRIAL INFORMATICS, IEEE/ASME TRANSACTIONS ON MECHATRONICS, IEEE TRANSACTIONS ON CIRCUITS AND SYSTEMS-I: REGULAR PAPERS, IEEE TRANSACTIONS ON COGNITIVE AND DEVELOPMENTAL SYSTEMS, IEEE TRANSACTIONS ON EMERGING TOPICS IN COMPUTATIONAL INTELLIGENCE, IEEE SYSTEMS JOURNAL, Engineering Applications of Artificial Intelligence (IFAC journal),
Science China Information Sciences, Science China Technological Sciences and Acta Automatica Sinica. He has been awarded as best/outstanding Associate Editor in IEEE journals for six times. He is a (leading) guest editor for several special issues focusing on autonomous systems, robotics, and industrial intelligence in IEEE TRANSACTIONS.
\end{IEEEbiography}

\clearpage
\onecolumn
\begin{appendices}
\renewcommand{\thetable}{A.\arabic{table}}
\setcounter{table}{0}
\section{Dataset description}\label{dataset_des}
We conduct spatio-temporal forecasting tasks on weather and traffic systems to predict temperature and traffic flows.\par
\textbf{\textit{Weather dataset.}}
For temperature forecasting, we use the ERA5 hourly dataset \cite{era5}, originally with a resolution of $\SI{0.25}{\degree}$, which we resample to $\SI{1}{\degree}$. Our study focuses on an area between $\SI{23}{\degree}$N and $\SI{35}{\degree}$N latitude and $\SI{100}{\degree}$E to $\SI{122}{\degree}$E longitude, encompassing 263 nodes. The dataset covers the period from January 1, 2012, to December 31, 2022, and uses historical 24-hour temperature data to predict temperatures 12 hours ahead.\par
\textbf{\textit{NYC datasets.}} 
We use traffic flow datasets for bikes and taxis in New York, which have been preprocessed by Ji \textit{et al.} \cite{ji2023spatio}. These datasets are segmented into three categories: NYCBike1, NYCBike2, and NYCTaxi, recording inflows and outflows of city bikes and taxis every 30 minutes. The NYCBike1 dataset spans from April 1st, 2014, to September 30th, 2014. The NYCBike2 dataset covers the period from July 1st, 2016, to August 29th, 2016, while the NYCTaxi dataset ranges from January 1st, 2015, to March 1st, 2015. The input and output setups are consistent with those described in Ji \textit{et al.} \cite{ji2023spatio}. In NYCBike1, we predict the inflow and outflow of 128 grids 30 minutes ahead using historical records of 9.5 hours, with a sequence length of 19. In NYCBike2 and NYCTaxi datasets, we utilize historical time series of 17.5 hours, with a sequence length of 35. The number of grids in NYCBike2 and NYCTaxi datasets is 200.\par
\textbf{\textit{PeMS04 and PeMS08 datasets.}}
The PeMS04 and PeMS08 datasets are subsets of the PeMS (PeMS Traffic Monitoring) dataset \cite{chen2001freeway}, which includes real-time traffic flow data collected from loop detectors on California highways. Specifically, PeMS04 contains traffic flow records from the San Francisco Bay Area, covering the period from January 1, 2018, to February 28, 2018. PeMS08 encompasses traffic data from July 1, 2016, to August 31, 2016. Both datasets are sampled at 5-minute intervals. In our forecasting task, we aim to predict traffic flow one hour ahead based on the past one-hour records. Both the input and output lengths for each prediction task are set to 12 time steps.\par
The training, validation, and test splits, along with the number of features, are summarized in Table~\ref{dataset_add}.\par
\section{\label{training}Training details of DRAN}
The training configuration of DRAN is as follows: 
The dimension of the adaptive node embedding is 80, consistent with STAEformer \cite{liu2023spatio}. 
In the SFL module, the Conv1d layers are configured with an input channel equal to the length of the lookback window $L$, an output channel of 1, a kernel size of 3, and ``circular padding'' as defined in the PyTorch package. The Linear layers map the feature dimension to a hidden dimension of 64.
The numbers of de-stationary attention and spatial attention layers
are both set to 3. Each de-stationary attention module follows the parameter setup of STAEformer \cite{liu2023spatio}, with 4 attention heads and a feed-forward dimension of 256. In the DSFL module, the feature dimensions of $\bm{H}_{\rm Dy}$ and $\bm{H}_{\rm St}$ are both set to 160, and the number of attention heads in the dynamic branch is set to 4.
In the Stochastic Learner, the latent layers include 3 Linear layers with ReLU activation functions, mapping the features to 64 dimensions. The reconstruction part of the Stochastic Learner consists of 3 Linear layers followed by ReLU activation functions, which remap the feature dimension from 64 to 160. The decoder comprises 2 Linear layers that fuse the deterministic and stochastic representations to produce the target feature outputs.
The balancing hyperparameters $\alpha$ and $\beta$ are fine-tuned experimentally to account for the stochastic nature and uncertainties of the datasets. Table \ref{hyper_select} presents the variation in prediction error of validation datasets corresponding to different values of $\alpha$ and $\beta$, with the value minimizing the error selected as optimal.\par
To quantify the dynamic variability of input signals for adaptive fusion, we compute two statistics from the input sequence $\bm{X}_{t-L:t}\in\mathbb{R}^{L\times N\times C}$, namely the variance ratio of first differences and the high-frequency energy ratio.
\begin{equation}
\bm{z}_{\rm dvr}
=
\frac{\mathrm{Var}(\Delta \bm{X})}{\mathrm{Var}(\bm{X})+\epsilon},
\end{equation}
\begin{equation}
\bm{z}_{\rm hf}
=
\frac{\sum_{f=k_c+1}^{F-1} P(f)}
{\sum_{f=0}^{F-1} P(f)+\epsilon},
\end{equation}
where $\Delta \bm{X}_{t}=\bm{X}_{t+1}-\bm{X}_{t}$, $P(f)$ denotes the temporal power spectrum of $\bm{X}$, $F=L//2+1$ is the number of frequency bins, $k_c=\lfloor 0.25(F-1)\rfloor$ is the cutoff index, and $\epsilon=10^{-6}$ is a small constant.
Finally, the two statistics are denoted as  $\bm{z}_{\rm dvr}$ and $\bm{z}_{\rm hf}$, respectively, and concatenated with the static features to construct the control feature for adaptive gating.\par
\begin{table*}[t]
\centering
\caption{Additional datasets details}
\footnotesize
\renewcommand{\arraystretch}{0.6}
\begin{tabular}{@{}c|cccccccccc@{}}
\toprule%
\textbf{Attributes}     & \textbf{\makecell{Training\\set} }
& \textbf{\makecell{Test\\set}} & \textbf{\makecell{Validation\\set}} & \textbf{\makecell{Node\\number}} & \textbf{\makecell{Feature\\number}}  \\ \midrule
Weather        & 5,623        & 1,607       & 803            & 263         & 1      \\ \midrule
NYCBike1   & 3,023        & 864         & 431            & 128         & 2            \\
NYCBike2   & 1,912        & 546         & 274            & 200         & 2       \\
NYCTaxi    & 1,912        & 546         & 274            & 200         & 2        \\ \midrule
PeMS04    & 10,181       & 3,394       & 3,394          & 307         & 1     \\
PeMS08     & 10,700       & 3,566       & 3,566          & 170         & 1   \\ \bottomrule
\end{tabular}
\label{dataset_add}
\end{table*}
\begin{table*}[!t]
\centering
\renewcommand{\arraystretch}{0.6} 
\footnotesize
\caption{The selection of balanced hyperparameters.}
\begin{tabular}{@{}cc|cc|cc|cc|cc|cc@{}}
\toprule
\multicolumn{2}{c|}{Weather} & \multicolumn{2}{c|}{NYCBike1} & \multicolumn{2}{c|}{NYCBike2} & \multicolumn{2}{c|}{NYCTaxi} & \multicolumn{2}{c|}{PeMS04} & \multicolumn{2}{c}{PeMS08} \\ \midrule
$\alpha$       & MAE         & $\alpha$      & MAE           & $\alpha$      & MAE           & $\alpha$       & MAE         & $\alpha$     & MAE          & $\alpha$     & MAE         \\ \midrule
0.001          & 0.856       & 0.001         & 5.102       & 0.001         & 4.958         & 0.001          & 11.375      & 0.001        & 18.186       & 0.001        & 13.604      \\
0.010          & 0.895       & 0.010         & 5.075         & 0.010         & 4.942         & 0.010          & 11.423      & 0.010        & 18.249       & 0.010        & 13.682      \\
0.050          & 0.667       & 0.050         & 5.037         & 0.050         & 4.855         & 0.050          & 11.191      & 0.050        & 18.343       & 0.050        & 13.560      \\
0.100          & 0.726       & 0.100         & 5.082         & 0.100         & 4.972         & 0.100          & 11.085      & 0.100        & 18.240       & 0.100        & 13.499      \\
0.500          & 0.758       & 0.500         & 5.101         & 0.500         & 5.040         & 0.500          & 10.892      & 0.500        & 18.299       & 0.500        & 13.526      \\
1.000          & 0.804       & 1.000         & 5.159         & 1.000         & 5.011         & 1.000          & 10.737      & 1.000        & 18.223       & 1.000        & 13.561      \\ \midrule
$\beta$        & MAE         & $\beta$       & MAE           & $\beta$       & MAE           & $\beta$        & MAE         & $\beta$      & MAE          & $\beta$      & MAE         \\ \midrule
0.001          & 0.902       & 0.001         & 5.129         & 0.001         & 5.030         & 0.001          & 11.264      & 0.001        & 18.361       & 0.001        & 13.517      \\
0.050          & 0.732       & 0.050         & 5.096         & 0.050         & 4.931         & 0.050          & 11.758      & 0.050        & 18.418       & 0.050        & 13.376      \\
0.100          & 0.711       & 0.100         & 5.074         & 0.100         & 4.906         & 0.100          & 10.914      & 0.100        & 18.327       & 0.100        & 13.472      \\
0.500          & 0.678       & 0.500         & 5.050         & 0.500         & 4.864      & 0.500          & 10.751      & 0.500        & 18.339       & 0.500        & 13.422      \\
1.000          & 0.900       & 1.000         & 5.134         & 1.000         & 4.938         & 1.000          & 10.861      & 1.000        & 18.230       & 1.000        & 13.596      \\
5.000          & 1.094            & 5.000         & 5.203         & 5.000         & 5.023         & 5.000          & 11.369      & 5.000        & 18.144       & 5.000        &  13.679           \\ \midrule
Selection ($\alpha$, $\beta$)    & (0.05, 0.5)   &($\alpha$, $\beta$)             &( 0.05, 0.5)     & ($\alpha$, $\beta$)        &(0.05, 0.5)     & ($\alpha$, $\beta$)               &(1, 0.5)        & ($\alpha$, $\beta$)             &(0.001, 5)     & ($\alpha$, $\beta$)       & (0.1, 0.05)    \\ \bottomrule
\end{tabular}
\label{hyper_select}
\end{table*}
\begin{table*}[!t]
\centering
\renewcommand{\arraystretch}{1} 
\footnotesize
\caption{\label{select}Hyperparameter candidate ranges}
\begin{tabular}{@{}c|l@{}}
\toprule
Methods    & Hyperparameters selection                                                                                  \\ \midrule
DA-RNN \cite{dualts}     & hidden dimension \{64, 128,256\}                                                                           \\ \hline
InfoTS \cite{infots}    & $\alpha$  \{0.01, 0.1, 0.5, 1, 5,   10\}; $beta$:  \{0.01, 0.1, 0.5, 1, 5,   10\}                          \\ \hline
AutoTCL \cite{parats}   & $\beta$ \{0.0003, 0.001, 0.01, 0.1,0.3\}; $\lambda$ \{0.0003, 0.001, 0.01,   0.1,0.3\}; local loss \{0.0003, 0.001, 0.01, 0.1, 0.5\}                      \\ \hline
TGCN \cite{zhao2019t}      & recurrent layer \{1, 2, 3\}; hidden dimension \{32, 64, 128\}                                              \\ \hline
GCGRU \cite{seo2018structured}     & recurrent layer \{1, 2, 3\}; hidden dimension \{32, 64, 128\}                                              \\ \hline
AGCRN \cite{bai2020adaptive}     & recurrent layer \{1, 2, 3\}; hidden dimension \{32, 64, 128\}                                              \\ \hline
DCRNN \cite{li2018diffusion}     & recurrent layer \{1, 2, 3\}; hidden dimension \{32, 64, 128\}                                              \\ \hline
STGCN \cite{yu2018spatio}     & spatio-temporal block \{1, 2, 3\}; hidden dimension \{32, 64, 128\}                                        \\ \hline
ASTGCN \cite{guo2019attention}    & spatio-temporal block \{1, 2, 3\}; hidden dimension \{32, 64, 128\}                                        \\ \hline
MegaCRN \cite{Meta_Graph}    & AGCRN layer \{1, 2, 3\}; memory dimension \{32, 64, 96\}; memory number   \{20, 40, 60\}                   \\ \hline
STAEformer \cite{liu2023spatio} & adaptive embedding dimension \{64, 80, 128\}                                                               \\ \hline
ST-SSL \cite{ji2023spatio}    & $\tau$ \{0.01, 0.1, 0.5, 1\}; cluster number \{4, 6, 10, 15, 20\}                                             \\ \hline
RGSL \cite{ijcai2022p328}      & spatio-temporal layer \{1, 2, 3\}; hidden dimension \{32, 64, 128\}                                        \\ \hline
TESTAM \cite{lee2024testam}     & quantile loss ratio \{0.1, 0.2, 0.3, 0.4, 0.5, 0.6, 0.7, 0.8, 0.9, 1\};   hidden dimension \{32, 64, 128\} \\ \hline
MemDA \cite{memda}     & memory units \{20, 64, 96\}; memory dimension \{32, 64, 128\}; $k$ \{5,   10, 15\}                         \\ \hline
STDN \cite{stdn} & decoder layer \{1,2,3\}; hidden dimension \{32, 64,128\} \\ \hline
DST-Mamba \cite{dstmamba} & Mamba layer \{1,2,3\}; Mamba state dimension \{32,64\}; channel expand \{1,2,3\} \\
\bottomrule
\end{tabular}
\end{table*}
\begin{table*}[htbp]
\centering
\footnotesize
\renewcommand{\arraystretch}{0.6} 
\caption{\label{ft}Final tuned hyperparameters}
\begin{tabular}{@{}c|c|cccccc@{}}
\toprule
\multirow{2}{*}{Methods} & \multirow{2}{*}{Parameter settings}                                          & \multirow{2}{*}{Weather} & \multirow{2}{*}{NYCBike1} & \multirow{2}{*}{NYCBike2 } & \multirow{2}{*}{NYCTaxi} & \multirow{2}{*}{PeMS04} & \multirow{2}{*}{PeMS08} \\
                         &                                                                              &                              &                           &                           &                          &                          &                         \\ \midrule
DA-RNN \cite{dualts}                  & hidden dimension                                                             & 64                           & 64                        & 128                       & 128                      & 128                      & 128                     \\ \midrule
\multirow{2}{*}{InfoTS \cite{infots}}  & $\alpha$                                                                     & 0.5                          & 5                         & 10                        & 5                        & 5                        & 1                       \\
                         & $\beta$                                                                      & 10                           & 0.5                       & 10                        & 0.5                      & 0.5                      & 5                       \\ \midrule
\multirow{3}{*}{AutoTCL \cite{parats}} & $\beta$                                                                      & 0.01                         & 0.1                       & 0.1                       & 0.1                      & 0.3                      & 0.3                     \\
                         & $\lambda$                                                                    & 0.1                          & 0.01                      & 0.001                     & 0.01                     & 0.3                      & 0.1                     \\
                         & Local loss                                                                   & 0.1                          & 0.0003                    & 0.0003                    & 0.0003                   & 0.01                     & 0.001                   \\ \midrule
\multirow{2}{*}{TGCN \cite{zhao2019t}}    & layer number                                                                 & 2                            & 2                         & 3                         & 2                        & 2                        & 3                       \\
                         & hidden dimension                                                             & 32                           & 32                        & 32                        & 32                       & 32                       & 64                      \\ \midrule
\multirow{2}{*}{GCGRU \cite{seo2018structured}}   & layer number                                                                 & 2                            & 2                         & 3                         & 2                        & 3                        & 3                       \\
                         & hidden dimension                                                             & 32                           & 32                        & 64                        & 32                       & 64                       & 64                      \\ \midrule
\multirow{2}{*}{AGCRN \cite{bai2020adaptive}}   & layer number                                                                 & 1                            & 1                         & 1                         & 1                        & 2                        & 1                       \\
                         & hidden dimension                                                             & 32                           & 64                        & 64                        & 64                       & 32                       & 64                      \\ \midrule
\multirow{2}{*}{DCRNN \cite{li2018diffusion}}   & layer number                                                                 & 2                            & 2                         & 2                         & 2                        & 2                        & 2                       \\
                         & hidden dimension                                                             & 32                           & 32                        & 64                        & 64                       & 32                       & 32                      \\ \midrule
\multirow{2}{*}{STGCN \cite{yu2018spatio}}   & layer number                                                                 & 2                            & 2                         & 2                         & 2                        & 2                        & 2                       \\
                         & hidden dimension       & 32    & 32     & 32    & 32      & 32         & 32       \\ \midrule
\multirow{2}{*}{ASTGCN \cite{guo2019attention}}  & layer number                                                                 & 2                            & 1                         & 2                         & 2                        & 2                        & 2                       \\
                         & hidden dimension                                                             & 64                           & 64                        & 128                       & 64                       & 64                       & 128                     \\ \midrule
\multirow{3}{*}{MegaCRN \cite{Meta_Graph}} & layer number                                                                 & 2                            & 2                         & 2                         & 2                        & 2                        & 2                       \\
                         & memory number                                                                & 20                           & 20                        & 60                        & 60                       & 20                       & 20                      \\
                         & memory dimension                                                             & 32                           & 32                        & 64                        & 64                       & 32                       & 32                      \\ \midrule
STAEformer               & adaptive dimension                                                           & 80                           & 80                        & 80                        & 80                       & 80                       & 80                      \\ \midrule
\multirow{2}{*}{ST-SSL \cite{ji2023spatio}}  & $\tau$                                                                       & 0.1                          & 0.5                       & 0.5                       & 0.5                      & 0.01                     & 0.1                     \\
                         & cluster number                                                               & 10                           & 6                         & 10                        & 4                        & 10                       & 15                      \\ \midrule
\multirow{2}{*}{RGSL \cite{ijcai2022p328}}    & layer number                                                                 & 2                            & 2                         & 3                         & 3                        & 2                        & 2                       \\
                         & hidden dimension                                                             & 64                           & 64                        & 64                        & 128                      & 64                       & 64                      \\ \midrule
\multirow{2}{*}{TESTAM \cite{lee2024testam}}  & loss                                                                         & 0.7                          & 0.7                       & 0.8                       & 0.9                      & 0.7                      & 0.7                     \\
                         & hidden dimension                                                             & 32                           & 32                        & 32                        & 32                       & 32                       & 32                      \\ \midrule
MemDA \cite{memda}                   & memory number                                                                & 20                           & 20                        & 40                        & 40                       & 20                       & 20                      \\
                         & memory dimension                                                             & 32                           & 32                        & 128                       & 64                       & 32                       & 32                      \\
                         & k of NTN & 5                            & 5                         & 15                        & 10                       & 5                        & 5                       \\ \midrule
STDN \cite{stdn} & layer num   &3 &3 &3 &3 &2 &3 \\
                 & hidden dimension &32 &32 &64 &32 &32 &64 \\ \midrule
DST-Mamba \cite{dstmamba} & Mamba layer    &2 &3 &3 &3 &2 &1 \\
                        & state dimension &32 &64 &64 &64 &32 &32 \\
                        &channel expand &1 &2 &1 &2 &1 & 1 \\     \bottomrule
\end{tabular}
\end{table*}
\section{Implementation Details}
\subsection{\label{baseline_app}Baselines and Implementation Details}
We compare our method against several baseline approaches, including state-of-the-art multivariate time series forecasting methods and spatio-temporal forecasting methods. Spatio-temporal forecasting techniques can be classified into two categories based on how they learn spatial relations: task-adaptive and dynamic-adaptive methods. Task-adaptive methods focus on learning static temporal and spatial relations from training datasets, which remain fixed during the testing phase. In contrast, dynamic-adaptive methods capture dynamically changing relations from input windows or by updating memory with incoming data.\par
In multivariable time series forecasting, DA-RNN \cite{dualts} is a classic dual-stage attention-based recurrent neural network designed to capture long-term temporal dependencies. InfoTS \cite{infots} and AutoTCL \cite{parats} focus on enhancing time series representation learning through series augmentations and contrastive learning. InfoTS introduces a novel contrastive learning approach with information-aware augmentations that adaptively select optimal augmentations and a meta-learner network to learn from datasets. AutoTCL achieves unified and meaningful time series augmentations at both the dataset and instance levels, leveraging information theory to enhance representation quality.\par
For spatio-temporal forecasting methods, TGCN \cite{zhao2019t}, STGCN \cite{yu2018spatio}, DCRNN \cite{li2018diffusion} and GCGRU \cite{seo2018structured} utilize physical spatial relations as adjacency matrices and employ static neural networks for prediction. TGCN, DCRNN, and GCGRU use Graph Convolutional Networks (GCN) and Recurrent Neural Networks (RNN) to capture spatial and temporal features. STGCN combines temporal and graph convolutions to learn spatial and temporal dependencies. AGCRN \cite{bai2020adaptive} utilizes learnable node embeddings to adapt spatial relations to tasks and node-adaptive parameters to capture specific attributes of each node. ASTGCN \cite{guo2019attention} and STAEformer \cite{liu2023spatio} employ attention mechanisms to capture dynamic changes in input features. RGSL \cite{ijcai2022p328} learns spatio-temporal dependencies from a predefined graph and learnable node embeddings, dynamically fusing features from two graphs using an attention mechanism. Additionally, STAEformer employs learnable node embeddings and concatenates them with input time series to capture static features. MegaCRN \cite{Meta_Graph} utilizes node embeddings to learn the static relations and memory networks to dynamically match sample patterns with learned static features. ST-SSL \cite{ji2023spatio} does not learn static relations and only uses the adjacency matrix based on node distances as a prior graph. It fine-tunes the static graph using node similarities, effectively fusing dynamic and static features. TESTAM \cite{lee2024testam} employs a mixture-of-experts model with three experts: one for temporal modeling, one for spatio-temporal modeling with a static graph, and one for spatio-temporal dependency modeling with a dynamic graph. MemDA \cite{memda} utilizes meta-dynamic network and memory to encode concept drift and adjust on the fly. DST-Mamba \cite{dstmamba} decomposes the temporal patterns of nodes into trend and seasonal components, and captures seasonal variations in a node-centric manner using Mamba modules. STDN \cite{stdn} employs a dynamic graph with spatio-temporal embeddings to learn global representations, while disentangling the trend and seasonal components of node signals. \par
We also fine-tune the hyperparameters of baseline methods to ensure fairness, especially for datasets not included in their official implementations. The candidate ranges for hyperparameter selection are shown in Table \ref{select}, and the final tuned values are summarized in Table \ref{ft}.\par
For the inference time comparison experiments, we use input data of the same size across all models and repeat the inference process 100 times to compute the average inference time per batch on the NYCTaxi dataset. All experiments are conducted on an NVIDIA RTX 3090 GPU.
\subsection{Implementation Details of Normalization Variants}
\label{app:norm_impl}
To ensure a controlled comparison, all normalization variants use the 
same data splits, input and forecasting lengths, model dimensions, 
optimizer, training epochs, and random seeds as DRAN. The temporal 
normalization component is the only component replaced, while the 
temporal learner, DSFL module, stochastic learner, and decoder remain 
unchanged. For each paired comparison, the variant with SFL differs 
from its counterpart only by the inclusion of SFL and 
$\mathcal{L}_{\rm spa}$.

For DAIN, Non-stationary Transformer, and Dish-TS, we implement their 
normalization operations following the corresponding original methods. 
The resulting temporally processed representations are passed through 
the shared temporal backbone, after which SFL is applied before spatial 
dependency modeling.

For ST-Norm, temporal normalization and spatial normalization are 
applied to the input separately. Following the original design, the 
temporally normalized features, spatially normalized features, and raw 
input features are concatenated and projected to the common latent 
dimension. The variants with and without SFL share the same feature 
composition and downstream architecture; 
SFL is applied to the fused representation before DSFL.
\subsection{Robustness Evaluation Details}
\label{app:robustness}
We evaluate robustness on the Weather and PeMS08 datasets,
representing weather and traffic forecasting tasks, respectively.
All models are trained on the original clean training data and directly
evaluated under different test conditions without retraining.

Let $\bm{X}\in\mathbb{R}^{B\times L\times N\times C}$ denote an input
batch, where the first channel $\bm{X}_{:,:,:,0}$ contains the observed
variable and the remaining channels contain timestamp features.
All test-time corruptions are applied only to the observed-variable
channel, while the timestamp features remain unchanged.

\textbf{Gaussian sensor noise.}
For Gaussian corruption, independent noise is added to every observed
value:
\begin{equation}
    \widetilde{X}_{b,l,n,0}
    =
    X_{b,l,n,0}+\epsilon_{b,l,n},
    \qquad
    \epsilon_{b,l,n}\sim\mathcal{N}(0,\sigma^2),
\end{equation}
where $\sigma\in\{0.1,0.2,0.3\}$ controls the corruption severity.
Because corruption is applied after input standardization, $\sigma$ is
measured in the normalized observation space.

\textbf{Missing sensors.}
For a missing ratio $\rho\in\{10\%,20\%,30\%\}$, we randomly select
\begin{equation}
    N_{\rm miss}=\operatorname{round}(\rho N)
\end{equation}
nodes and construct a binary sensor mask
$\bm{m}\in\{0,1\}^{N}$, where $m_n=0$ denotes a missing sensor.
The corrupted input is computed as
\begin{equation}
    \widetilde{X}_{b,l,n,0}
    =
    m_n X_{b,l,n,0}.
\end{equation}
Within each evaluation run, the same sensor mask is reused across all
samples, batches, and lookback time steps. Thus, a selected node is
unavailable throughout the complete lookback window, simulating a
persistent sensor outage rather than independently missing
observations. Timestamp features are retained for all nodes.

\textbf{Naturally high-dynamic periods.}
In addition to synthetic corruption, we evaluate performance on
naturally high-dynamic test samples. For each test sample, its dynamic
score is computed from the forecasting target $\bm{Y}$ as
\begin{equation}
    s_b
    =
    \frac{1}{(H-1)NC}
    \sum_{h=1}^{H-1}
    \sum_{n=1}^{N}
    \sum_{c=1}^{C}
    \left|
        Y_{b,h+1,n,c}-Y_{b,h,n,c}
    \right|.
\end{equation}
We select the top $10\%$ of test samples according to $s_b$ and report
model performance on this subset. The forecasting targets are used
only for post-hoc subset construction and are not available to the
model during prediction.

For all conditions, robustness is evaluated using MAE over five random
seeds. The reported values are the corresponding mean and standard
deviation.
\section{\label{metrics}Evaluation Metrics}

Let
$\widehat{\bm{X}},\bm{X}
\in\mathbb{R}^{m\times H\times N\times C}$
denote the predicted and ground-truth sequences, respectively,
where $m$ is the number of test samples, $H$ is the forecasting
horizon, $N$ is the number of nodes, and $C$ is the number of
variables. All reported metrics are averaged over the corresponding
test dimensions, and lower values indicate better performance.

\noindent\textbf{Point-wise forecasting metrics.}
We use Mean Absolute Error (MAE) and Mean Absolute Percentage
Error (MAPE) to evaluate time-aligned prediction accuracy:
\begin{align}
\operatorname{MAE}
&=
\frac{1}{mHNC}
\sum_{s=1}^{m}
\sum_{h=1}^{H}
\sum_{i=1}^{N}
\sum_{c=1}^{C}
\left|
\widehat{X}_{s,h,i,c}
-
X_{s,h,i,c}
\right|,
\label{eq:mae}
\\
\operatorname{MAPE}
&=
\frac{100\%}{mHNC}
\sum_{s=1}^{m}
\sum_{h=1}^{H}
\sum_{i=1}^{N}
\sum_{c=1}^{C}
\left|
\frac{
\widehat{X}_{s,h,i,c}
-
X_{s,h,i,c}
}{
X_{s,h,i,c}
}
\right|.
\label{eq:mape}
\end{align}

\noindent\textbf{Distribution-level forecasting metric.}
We further use the empirical Wasserstein-1 Distance (WD) to
measure the discrepancy between the predicted and ground-truth
distributions. Unlike MAE and MAPE, WD does not require
sample-wise correspondence after the empirical distributions are
constructed.

For each forecasting step $h$, node $i$, and variable $c$, let
$\widehat{X}_{(s),h,i,c}$ and $X_{(s),h,i,c}$ denote the
$s$-th ordered values of
$\{\widehat{X}_{s,h,i,c}\}_{s=1}^{m}$ and
$\{X_{s,h,i,c}\}_{s=1}^{m}$, respectively.
The empirical Wasserstein-1 distance is
\begin{equation}
\operatorname{WD}_{h,i,c}
=
\frac{1}{m}
\sum_{s=1}^{m}
\left|
\widehat{X}_{(s),h,i,c}
-
X_{(s),h,i,c}
\right|.
\label{eq:wd_step}
\end{equation}

The overall WD is obtained by averaging over all forecasting
steps, nodes, and variables:
\begin{equation}
\operatorname{WD}
=
\frac{1}{HNC}
\sum_{h=1}^{H}
\sum_{i=1}^{N}
\sum_{c=1}^{C}
\operatorname{WD}_{h,i,c}.
\label{eq:wd}
\end{equation}
Thus, MAE and MAPE quantify point-wise forecasting errors,
whereas WD evaluates distributional alignment over the test set.\par
\noindent\textbf{Statistical significance test.}
To determine whether the performance differences between DRAN
and the strongest competing baseline are statistically significant,
we conduct a two-sided paired $t$-test for each dataset and
evaluation metric. Both methods are evaluated using the same
random seeds $\{31,32,33,34,35\}$.

For a given metric, let $a_r$ and $b_r$ denote the results of
DRAN and the competing baseline under the $r$-th valid seed,
respectively. Their paired difference is
\begin{equation}
d_r=a_r-b_r.
\label{eq:paired_diff}
\end{equation}

The mean and sample standard deviation of the paired differences
are computed as
\begin{align}
\bar{d}
&=
\frac{1}{n}
\sum_{r=1}^{n} d_r,
\\
s_d
&=
\sqrt{
\frac{1}{n-1}
\sum_{r=1}^{n}
(d_r-\bar{d})^2
},
\end{align}
where $n$ is the number of valid seed pairs and equals five when
all seed results are available. The test statistic is
\begin{equation}
t
=
\frac{\bar{d}}
{s_d/\sqrt{n}}.
\label{eq:t_test}
\end{equation}

Under the null hypothesis
\begin{equation}
H_0:\mathbb{E}[d_r]=0,
\end{equation}
the statistic follows a Student's $t$-distribution with $n-1$
degrees of freedom. We regard $p<0.05$ as statistically
significant and mark the corresponding result with $\dagger$.
Values smaller than $0.001$ are reported as $<0.001$.
The tests are implemented using
\texttt{scipy.stats.ttest\_rel} with
\texttt{alternative="two-sided"}. If either method lacks a result
for a particular seed, that seed pair is excluded from the test.

\noindent\textbf{Spatial-relation preservation metrics.}
For the spatial-distribution preservation analysis, we use
Pairwise Distance Distortion (PDD) and relation-level
Wasserstein-1 Distance $\operatorname{WD}_{\rm rel}$.

Let
$\bm{Z}\in
\mathbb{R}^{m\times L\times N\times D_Z}$
denote a representation to be evaluated, where $D_Z$ is its
feature dimension. Let
$\bm{Z}^{\rm ref}\in
\mathbb{R}^{m\times L\times N\times D_{\rm ref}}$
denote the corresponding reference representation, such as the
representation extracted from the raw input.

For each sample $s$ and time step $l$, we construct a spatial
relation matrix
$\bm{R}^{Z}_{s,l}\in\mathbb{R}^{N\times N}$
using the mean $\ell_1$ distance between node representations:
\begin{equation}
R^{Z}_{s,l,i,j}
=
\frac{1}{D_Z}
\left\|
Z_{s,l,i,:}
-
Z_{s,l,j,:}
\right\|_1.
\label{eq:relation_matrix}
\end{equation}
The reference relation matrix
$\bm{R}^{\rm ref}_{s,l}$
is constructed analogously from $\bm{Z}^{\rm ref}$.
Dividing by the feature dimension reduces the direct dependence
of the relation magnitude on the latent dimensionality.

PDD measures the mean absolute distortion of individual
off-diagonal node-pair relations:
\begin{equation}
\begin{aligned}
\operatorname{PDD}
\left(
\bm{Z},\bm{Z}^{\rm ref}
\right)
=
\frac{1}{mLN(N-1)}
\sum_{s=1}^{m}
\sum_{l=1}^{L}
\sum_{\substack{i,j=1\\i\neq j}}^{N}
\left|
R^{Z}_{s,l,i,j}
-
R^{\rm ref}_{s,l,i,j}
\right|.
\end{aligned}
\label{eq:pdd}
\end{equation}

To evaluate the distributional discrepancy between spatial
relations, we extract the
$M=N(N-1)$ off-diagonal entries of each relation matrix.
Let
$\bm{r}^{Z}_{s,l},\bm{r}^{\rm ref}_{s,l}\in\mathbb{R}^{M}$
contain these entries, and let
$r^{Z}_{s,l,(q)}$ and $r^{\rm ref}_{s,l,(q)}$
denote their $q$-th ordered values.
The relation-level Wasserstein-1 distance is
\begin{equation}
\begin{aligned}
\operatorname{WD}_{\rm rel}
\left(
\bm{Z},\bm{Z}^{\rm ref}
\right)
=
\frac{1}{mLM}
\sum_{s=1}^{m}
\sum_{l=1}^{L}
\sum_{q=1}^{M}
\left|
r^{Z}_{s,l,(q)}
-
r^{\rm ref}_{s,l,(q)}
\right|.
\end{aligned}
\label{eq:wd_rel}
\end{equation}

Lower PDD indicates smaller distortion of individual node-pair
relations, while lower $\operatorname{WD}_{\rm rel}$ indicates
better preservation of the overall spatial-relation distribution.

\section{\label{theory}Theoretical Analysis of SFL}

In this appendix, we provide the theoretical justification for the 
inequalities presented in the main text, which explain why the SFL 
module improves both node-wise and neighbor-wise spatial consistency.

Let $H_{\text{raw}}$ denote the 
un-normalized latent representations, and let 
$H_{\text{tem}}$ denote the temporally normalized representations.
The SFL module applies learned node-wise scaling and shifting to 
$H_{\text{tem}}$ to produce $H_{\text{spa}}$.

During training, SFL is optimized with the following reconstruction loss:
\begin{equation}
\mathcal{L}_{\rm spa}
=
\mathbb{E}_{t,i}
\left[
\left\|
\bm{H}_{{\rm spa},t,i,:}
-
\bm{H}_{{\rm raw},t,i,:}
\right\|^2_2
\right].
\end{equation}
where the expectation is taken over all nodes (and batches/time steps).
Note that $H_{\text{tem}}$ is always a feasible choice of $H_{\text{spa}}$,
i.e., $H_{\text{spa}} = H_{\text{tem}}$ corresponds to a valid 
parameter configuration.\par
\textbf{Lemma 1.}
Let $\bm{H}^{*}_{\rm spa}$ be a global minimizer of
$\mathcal{L}_{\rm spa}$ in Eq.~(4). If $\bm{H}_{\rm tem}$ is a
feasible output of SFL, then
\begin{equation}
\mathbb{E}_{t,i}
\left[
\left\|
\bm{H}^{*}_{{\rm spa},t,i,:}
-
\bm{H}_{{\rm raw},t,i,:}
\right\|_2^2
\right]
\leq
\mathbb{E}_{t,i}
\left[
\left\|
\bm{H}_{{\rm tem},t,i,:}
-
\bm{H}_{{\rm raw},t,i,:}
\right\|_2^2
\right].
\label{eq:lemma_node}
\end{equation}

\textit{Proof.}
Since $\bm{H}^{*}_{\rm spa}$ globally minimizes
$\mathcal{L}_{\rm spa}$ and $\bm{H}_{\rm tem}$ is a feasible
candidate, evaluating the objective at
$\bm{H}^{*}_{\rm spa}$ cannot yield a larger value than
evaluating it at $\bm{H}_{\rm tem}$. Therefore,
Eq.~\eqref{eq:lemma_node} follows directly.
\hfill$\square$
\par
For compactness, define
\begin{equation}
d(\bm{a},\bm{b})
=
\left\|
\bm{a}-\bm{b}
\right\|_2
\end{equation}
and
\begin{equation}
D_{\rm pair}(\bm{H})
=
\mathbb{E}_{t,i,j}
\left[
\left|
d(\bm{H}_{t,i,:},\bm{H}_{t,j,:})
-
d(\bm{H}_{{\rm raw},t,i,:},
  \bm{H}_{{\rm raw},t,j,:})
\right|
\right],
\label{eq:pair_definition}
\end{equation}
where the expectation is taken over node pairs $i\neq j$,
with the batch dimension omitted for notational simplicity.

\textbf{Lemma 2.}
The pairwise distance distortion of the optimal spatially
restored representation is bounded by
\begin{align}
D_{\rm pair}(\bm{H}^{*}_{\rm spa})
&\leq
2
\sqrt{
\mathbb{E}_{t,i}
\left[
\left\|
\bm{H}^{*}_{{\rm spa},t,i,:}
-
\bm{H}_{{\rm raw},t,i,:}
\right\|_2^2
\right]
}
\nonumber\\
&\leq
2
\sqrt{
\mathbb{E}_{t,i}
\left[
\left\|
\bm{H}_{{\rm tem},t,i,:}
-
\bm{H}_{{\rm raw},t,i,:}
\right\|_2^2
\right]
}.
\label{eq:pair_bound}
\end{align}

\textit{Proof.}
For each sample $b$, time step $t$, and node pair $(i,j)$,
the reverse triangle inequality gives
\begin{align}
&
\left|
d(
\bm{H}^{*}_{{\rm spa},b,t,i,:},
\bm{H}^{*}_{{\rm spa},b,t,j,:})
-
d(
\bm{H}_{{\rm raw},b,t,i,:},
\bm{H}_{{\rm raw},b,t,j,:})
\right|
\nonumber\\
&\leq
\left\|
\left(
\bm{H}^{*}_{{\rm spa},b,t,i,:}
-
\bm{H}^{*}_{{\rm spa},b,t,j,:}
\right)
-
\left(
\bm{H}_{{\rm raw},b,t,i,:}
-
\bm{H}_{{\rm raw},b,t,j,:}
\right)
\right\|_2
\nonumber\\
&\leq
\left\|
\bm{H}^{*}_{{\rm spa},b,t,i,:}
-
\bm{H}_{{\rm raw},b,t,i,:}
\right\|_2
+
\left\|
\bm{H}^{*}_{{\rm spa},b,t,j,:}
-
\bm{H}_{{\rm raw},b,t,j,:}
\right\|_2 .
\label{eq:reverse_triangle}
\end{align}

Taking the expectation over samples, time steps, and node
pairs yields
\begin{align}
D_{\rm pair}(\bm{H}^{*}_{\rm spa})
&\leq
2\,
\mathbb{E}_{t,i}
\left[
\left\|
\bm{H}^{*}_{{\rm spa},t,i,:}
-
\bm{H}_{{\rm raw},t,i,:}
\right\|_2
\right]
\nonumber\\
&\leq
2
\sqrt{
\mathbb{E}_{t,i}
\left[
\left\|
\bm{H}^{*}_{{\rm spa},t,i,:}
-
\bm{H}_{{\rm raw},t,i,:}
\right\|_2^2
\right]
},
\end{align}
where the second inequality follows from the
Cauchy--Schwarz inequality. Applying Lemma~1 and the
monotonicity of the square-root function gives the second
inequality in Eq.~\eqref{eq:pair_bound}.
\hfill$\square$

\newpage
\section{Additional Temporal Prediction Visualizations}
\label{app:temporal_vis}
\begin{figure}[!h]
\centering
\tiny
\includegraphics[width=0.5\linewidth]{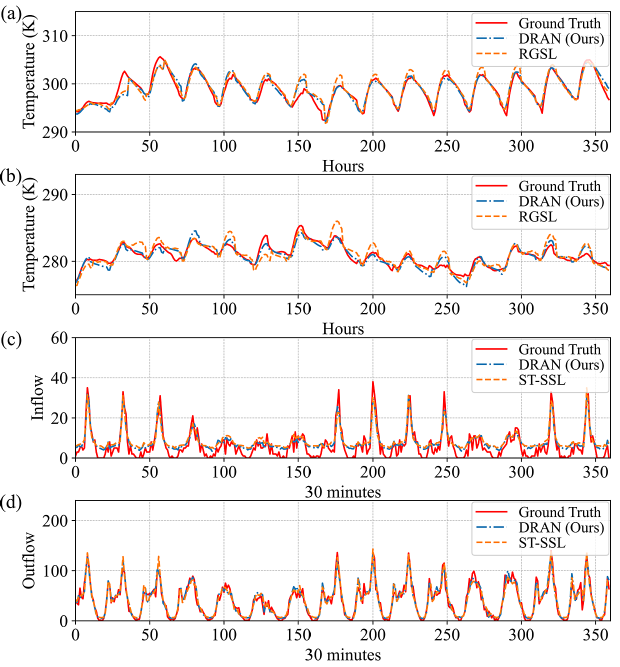}
\caption{\label{ts}Additional node-level temporal prediction results.
Panels (a) and (b) show temperature forecasts for two Weather
nodes, while panels (c) and (d) show inflow and outflow forecasts
for one NYCBike1 node.}
\end{figure}

In Fig. \ref{ts}, we visualize node-level time series of weather and NYCBike1 datasets. Both RGSL and DRAN capture overall trends and regular periodicity in the weather dataset, though they are less accurate at some extreme values. In Fig. \ref{ts} (c) and (d), DRAN more accurately predicts sudden drops in traffic flow. \par
\newpage
\end{appendices}

\end{document}